\documentclass[a4paper, 10pt]{article}

\usepackage[english]{babel}
\usepackage{graphicx}
\usepackage{amsmath}
\usepackage{amssymb}
\usepackage{tikz}
\usepackage{multirow}
\usepackage{url}
\usepackage{a4wide}

\usepackage{caption}
\usepackage[affil-it]{authblk}
\usepackage{etoolbox}

\usepackage[ruled,vlined]{algorithm2e}
\DontPrintSemicolon

\makeatletter
\def\cl@chapter{}
\makeatother
\usepackage{cleveref}
\crefname{}{}{}

\newcommand{\R}{\mathbb{R}}
\newcommand{\diff}{\mathop{}\!\mathrm{d}}
\newcommand{\dRPCA}{\delta\text{-RPCA}}

\SetCommentSty{mycommfont}

\makeatletter
\newcommand*\bigcdot{\mathpalette\bigcdot@{.5}}
\newcommand*\bigcdot@[2]{\mathbin{\vcenter{\hbox{\scalebox{#2}{$\m@th#1\bullet$}}}}}
\makeatother

\DeclareMathOperator{\vect}{vec}
\DeclareMathOperator{\argmin}{argmin}
\DeclareMathOperator{\TV}{TV}
\DeclareMathOperator{\id}{id}
\DeclareMathOperator{\diag}{diag}
\DeclareMathOperator{\rank}{rank}

\title{Deformable Groupwise Image Registration using Low-Rank and Sparse Decomposition}

\author[1]{Roland Haase\thanks{Corresponding author. Contact: \texttt{roland.haase@mic.uni-luebeck.de}}}
\author[2]{Stefan Heldmann}
\author[1]{Jan Lellmann}

\affil[1]{Institute of Mathematics and Image Computing, University of L\"ubeck}
\affil[2]{Fraunhofer MEVIS, L\"ubeck}

\date{}

\begin{document}
	
	\maketitle
	
	\begin{abstract}
		Low-rank and sparse decompositions and robust PCA (RPCA) are highly successful techniques in image processing and have recently found use in groupwise image registration.
		In this paper, we investigate the drawbacks of the most common RPCA-dissimi\-larity metric in image registration and derive an improved version.
		In particular, this new metric models low-rank requirements through explicit constraints instead of penalties and thus avoids the pitfalls of the established metric.
		Equipped with total variation regularization, we present a theoretically justified multilevel scheme based on first-order primal-dual optimization to solve the resulting non-parametric registration problem.
		As confirmed by numerical experiments, our metric especially lends itself to data involving recurring changes in object appearance and potential sparse perturbations.
		We numerically compare its peformance to a number of related approaches.
	\end{abstract}

	\begin{center}
	{\bf \small Keywords}\\[.25em]
	{\small Groupwise Image Registration $\diamond$ Motion Correction \\ Low-Rank/Sparse Decomposition $\diamond$ RPCA}
	\end{center}
	
	\section{Introduction}
	\label{sec:intro}
	
	\subsection{Groupwise Image Registration}
	\label{subsec:groupwise}
	The problem of aligning one image with another image of the same object is a well-studied problem in image processing and variational methods have proven successful for the task \cite{Modersitzki2003,Sotiras2013}.
	However, many application scenarios involve data comprised of more than two images, as in the case of image data gathered over time, which necessitates \textit{groupwise} methods.
	Naive pairwise techniques, that select one image from the group as a fixed reference and register all other images to the reference have been shown to be inconsistent with respect to registration accuracy (depending on the choice of the reference) and are generally deemed inferior to groupwise methods \cite{Metz2011,Huizinga2016}.
	These allow all images of the group to be deformed simultaneously and therefore operate on an \textit{implicit reference}.
	
	A crucial step in solving any image registration problem is the selection of a suitable dissimilarity metric on pairs or groups of images.
	In the past, both generalizations of established dissimilarity metrics for the classic two image problem and new concepts have been proposed to measure the distance between a group of $N > 2$ images.
	Examples for the former case include the \textit{variance}-measure found in \cite{Bhatia2007,Metz2011} that extends the well-known \textit{sum of squared distances}, different generalizations of the \textit{mutual information} from \cite{Polfliet2018,Huizinga2016} and a multi-image version of the \textit{normalized gradient fields}-measure in \cite{Brehmer2018}.
	
	One example of a newly developed metric that is also related to the metric proposed in this work is $D_{\text{PCA2}}$ from \cite{Huizinga2016}.
	Given $N$ images $T_1, \ldots, T_N \in \R^{m \times n}$, this measure operates on the so-called \textit{Casorati matrix}
	\begin{equation}
	M_{T_1, \ldots, T_N} := [ \vect(T_1), \ldots, \vect(T_N) ] \in \R^{m n \times N},
	\label{eq:casorati}
	\end{equation}
	where $\vect(\cdot)$ denotes a column-major vectorization.
	In $D_{\text{PCA2}}$, one proceeds to penalize a weighted sum of the (nonnegative) eigenvalues $\lambda_i$ of the correlation matrix
	\begin{equation}
	K := \frac{\Sigma^{-1} (M_{T_1, \ldots, T_N} - \bar{M})^{\top} (M_{T_1, \ldots, T_N} - \bar{M}) \Sigma^{-1}}{N - 1}.
	\label{eq:correlation}
	\end{equation}
	$\bar{M}$ is the repeated columnwise mean of $M_{T_1, \ldots, T_N}$ and $\Sigma$ is diagonal with diagonal elements given by the standard deviations of the columns of $M_{T_1, \ldots, T_N}$.
	To be exact, the metric is given by
	\begin{equation}
	D_{\text{PCA2}}(T_1, \ldots, T_N) := \sum_{i=1}^N i \lambda_i.
	\label{eq:dpca2}
	\end{equation}
	As the number of nonzero eigenvalues of $K$ is equal to the rank of $M_{T_1, \ldots, T_N}$, minimizing $D_{\text{PCA2}}$ promotes low-rankness of $M_{T_1, \ldots, T_N}$ and similarity between images is modeled as linear dependency.
	Note that apart from the $\Sigma^{-1}$-weighting in \eqref{eq:correlation}, the eigenvalues $\lambda_i$ correspond to variances along the principal components of $M_{T_1, \ldots, T_N}$, which emphasizes the relation to the eponymous PCA.
	$D_{\text{PCA2}}$ will serve as a comparison method for our proposed metric in the experiments of section~\ref{sec:results}.
	
	\subsection{Robust PCA}
	\label{subsec:rpca}
	As the classic PCA is known for its sensitivity towards sparsely distributed outliers, such methods are prone to fail for datasets involving partially unreliable data or strong changes in image intensity over time.
	To overcome this issue, different versions of a \textit{Robust PCA} (RPCA) were proposed in the literature -- see \cite{Guyon12} for an extensive comparison.
	The most widely-used RPCA-variant is arguably the \textit{Principal Component Pursuit} (PCP) from \cite{Chandrasekaran2011,Candes2011}.
	PCP is derived as a convex relaxation of the combinatorial optimization problem
	\begin{equation}
	\min_{L, E \in \R^{p \times q}} \rank(L) + ||E||_0 \quad \mbox{s.t. } M = L + E
	\label{eq:rpca_combinatorial}
	\end{equation}
	for given data $M \in \R^{p \times q}$.
	The term $||E||_0$ denotes the number of non-zero entries of $E$.
	Replacing both summands of \eqref{eq:rpca_combinatorial} with their convex hulls yields
	\begin{equation}
	\min_{L \in \R^{p \times q}} ||L||_* + ||M - L||_1,
	\label{eq:pcp}
	\end{equation}
	which is convex in $L$ and thus poses a more tractable optimization problem.
	$||L||_*$ is the so-called \textit{nuclear norm}, defined as the sum of all singular values of $L$ (see \cite{Golub1996}) and $|| M - L ||_1 = \sum_{i = 1}^p \sum_{j = 1}^q | M_{i, j} - L_{i, j} |$ is a $\ell_1$-type norm.
	Especially recall the relationship between the singular values $\sigma_i$ and the rank of a matrix: $\rank(A) = \#\{\sigma_i(A) > 0\}$ (see again \cite{Golub1996}).
	The decomposition of $M$ generated by \eqref{eq:pcp} is usually referred to as a \textit{low-rank and sparse decomposition}, in which $L$ is of low rank and $E = M - L$ is sparse.
	
	PCP has previously been used in the context of groupwise image registration by \cite{Peng2010,Heber2014,Hamy2014,Liu2014}.
	Primarily tackled therein were datasets for which low-dimensio\-nal approximations using PCA-based techniques were not applicable due to occlusions, local changes in image intensity (for the case of DCE-MRI data) and irregular pathologies in medical image data.
	In all these publications, the data matrix $M$ for \eqref{eq:pcp} was constructed as a Casorati matrix \eqref{eq:casorati}.
	The authors of \cite{Hamy2014,Liu2014} however only used low-rank and sparse decompositions as preprocessing steps and performed subsequent registrations on the generated low-rank components $L$ with different algorithms.
	Contrary to that, \cite{Peng2010,Heber2014} both used the optimal value of \eqref{eq:pcp} as a metric for the similarity of a set of given images~${T_1, \ldots, T_N}$.
	
	Section \ref{sec:rpca_distance} of this paper will present a deeper analysis of PCP as a distance measure.
	We argue that PCP has some inherent drawbacks:
	Perfect alignments of all $T_i$ often constitute local minimizers of PCP in only very narrow neighborhoods.
	At the same time, degenerated deformations result in comparatively lower energies.
	To overcome these issues, we present in this work a modification of PCP that is still convex and therefore easy to optimize.
	
	\subsection{Proposed Approach}
	Precisely, we propose to use the following groupwise dissimilarity measure:
	\begin{equation}
	D_{\dRPCA}(T_1, \ldots, T_N) := \min_{L \in \R^{m n \times N}} || M_{T_1, \ldots, T_N} - L ||_1 \quad \mbox{s.t. } || L - \bar{L} ||_* \leq \nu.
	\label{eq:mod_measure_intro}
	\end{equation}
	Here $M_{T_1, \ldots, T_N}$ is again the Casorati matrix \eqref{eq:casorati}, $\bar{L}$ is the repeated columnwise mean of $L$ and $\nu \geq 0$ is a suitable threshold for the nuclear norm.
	The intuition behind \eqref{eq:mod_measure_intro} is to jointly measure the $\ell_1$-distance between the input images and their optimal approximations in a low-dimen\-sio\-nal linear subspace.
	Details are given in section~\ref{sec:rpca_distance}.
	Our main contributions in this work involve the following:
	\begin{itemize}
		\item A novel technique for low-rank and sparse decompositions that results in a more suitable distance metric for groupwise registration tasks than previous approaches.
		\item A less restrictive uniqueness constraint than the one commonly employed in the literature.
		\item A multi-level strategy with theoretically justified scaling that solves the registration model in an iterative process and that uses first-order primal-dual optimization techniques to solve the subproblems.
	\end{itemize}
	
	\subsection{Other Related Work}
	\label{subsec:delineation}
	Major differences between our approach and related methods for variational groupwise registration are as follows:
	
	Besides the fact that the $D_{\text{PCA2}}$-measure from \cite{Huizinga2016} is based on the classic PCA (whereas ours is based on RPCA), the authors suggest a parametric deformation model based on B-Splines.
	Instead, we employ a non-parametric model that is fully deformable and that is explicitly (and flexibly) regularized through a total variation penalty (see section~\ref{sec:regularization}).
	Regularization in \cite{Huizinga2016} is handled implicitly through grid point spacing, and the same is true for all of \cite{Bhatia2004,Bhatia2007,Metz2011,Huizinga2016,Guyader2018,Polfliet2018}, as they use B-Spline deformations in the same manner.
	Concerning the two PCP-based registration approaches \cite{Peng2010,Heber2014}, the former is even further restricted to affine deformations, while the latter operates on light-field data, for which a geometric relationship between input images is known \textit{a priori} and is exploited in the registration process.
	
	Another non-parametric approach is presented in \cite{Brehmer2018}.
	While also based on rank minimization, the authors use normalized image gradients as feature vectors and define alignments locally (instead of image intensities as features and global alignments as in this article).
	A continuation of \cite{Brehmer2018} is found in \cite{Brehmer2019}, which generalizes the former approach to different kinds of feature vectors and formulates alignments globally.
	
	\subsection{Outline and Contributions}
	\label{subsec:outline}
	The remainder of this article is organized as follows:
	In section~\ref{sec:rpca_distance}, we analyze the established PCP-metric and derive our proposed approach as a replacement.
	In section~\ref{sec:regularization}, the total variation is discussed as a regularizer for our model and a new uniqueness constraint for groupwise image registration algorithms is introduced.
	In section~\ref{sec:optimization}, an in-depth account of the optimization strategy and its implementation is given, including a multilevel scheme with theoretically justified scaling.
	In the subsequent sections~\ref{sec:data} and~\ref{sec:results}, we introduce the benchmark data and present a numerical comparison to related approaches.
	Section~\ref{sec:conclusion} gives concluding remarks.
	
	\section{RPCA-based Distance Measures}
	\label{sec:rpca_distance}
	
	\subsection{Classical Approach}
	\label{subsec:naive_distance}
	The classical PCP image distance from \cite{Peng2010,Heber2014} is given by
	\begin{equation}
	D_{\text{PCP}} (T_1, \ldots, T_N) := \min_{L \in \R^{m n \times N}} || L ||_* + \mu || M - L ||_1
	\label{eq:pcp_measure}
	\end{equation}
	with $M$ as a Casorati matrix\footnote{From here on, we omit the explicit notation of the dependence of $M$ on $T_1, \ldots, T_N$ for readability.}.
	The parameter $\mu > 0$ controls the weighting between the requirement on $L$ to be of low rank and the requirement on ${E := M - L}$ to be sparse.
	
	\begin{figure}[t]
		\resizebox{\textwidth}{!}{
			\begin{tikzpicture}
			\node[inner sep=0pt] at (0, 0) {\includegraphics[width=2cm]{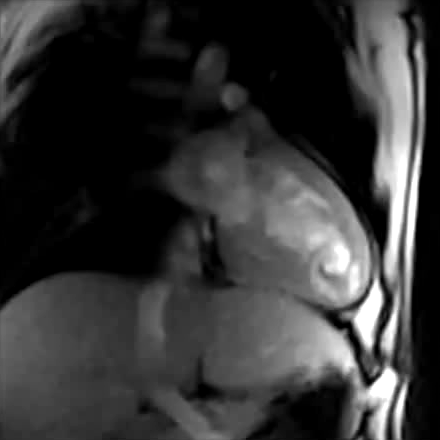}};
			\node[inner sep=0pt] at (2.5, 0) {\includegraphics[width=2cm]{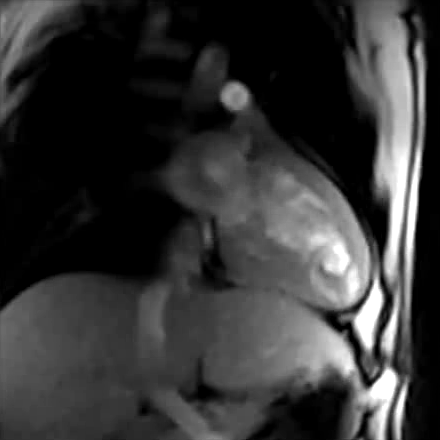}};
			\node[inner sep=0pt] at (5, 0) {\includegraphics[width=2cm]{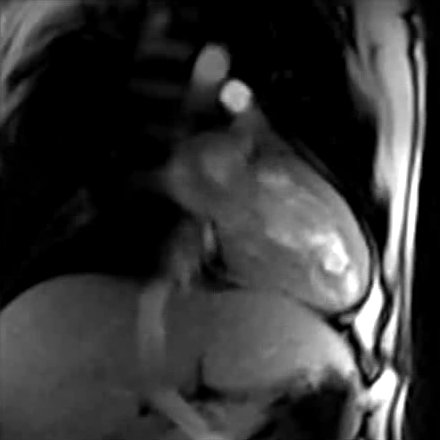}};
			\node[inner sep=0pt] at (7.5, 0) {\includegraphics[width=2cm]{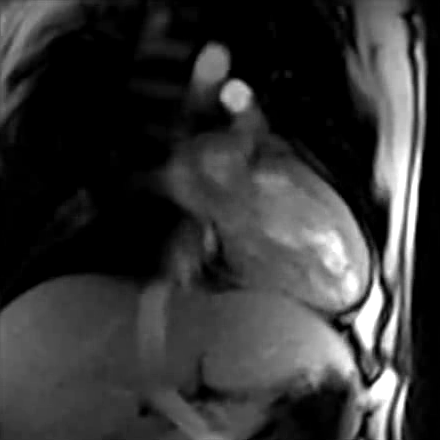}};
			\node[inner sep=0pt] at (10, 0) {\includegraphics[width=2cm]{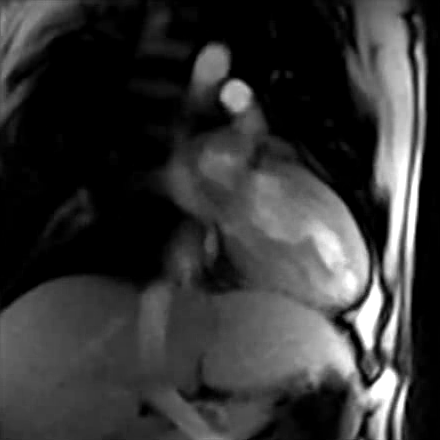}};
			\end{tikzpicture}
		}
		\caption{
			Depicted are five subsequent frames $T_1, \ldots, T_5$ from a MRI sequence that serve as input for the experiments in Fig.~\ref{fig:distance_measure_exp}:
			In order to analyze the behavior of a given dissimilarity measure, its energy is determined while one image is kept fixed and the remaining images are warped uniformly in a prescribed manner.
			Due to their short temporal offset, the input images $T_1, \ldots, T_5$ can be regarded as aligned
		}
		\label{fig:distance_measure_exp_frames}
	\end{figure}
	
	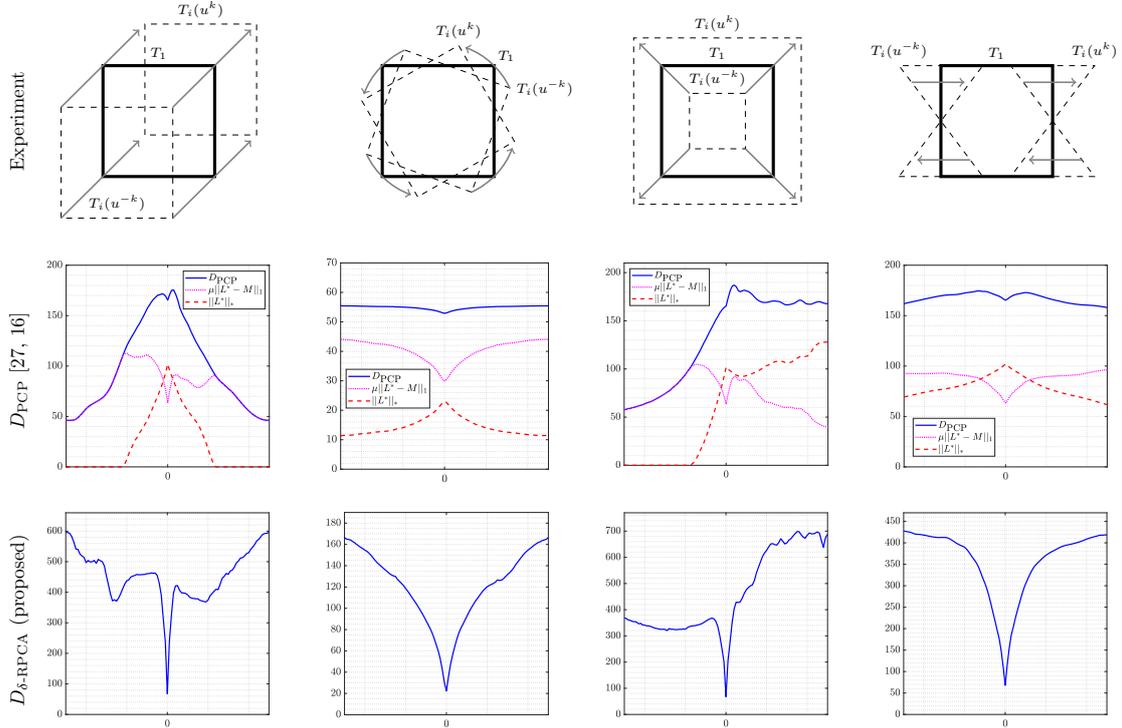
\begin{figure}[t]
		\resizebox{\textwidth}{!}{
			\begin{tikzpicture}
			
			\node[rotate=90] at (-1.5, 1) {Experiment};
			\node[rotate=90] at (-1.5, -3.5) {$D_{\text{PCP}}$ \cite{Peng2010,Heber2014}};
			\node[rotate=90] at (-1.5, -8) {$D_{\dRPCA}$ (proposed)};
			
			\draw[ultra thick] (0, 0) rectangle (2, 2);
			\draw (1, 2) node[above] {\scriptsize $T_1$};
			\draw[dashed] (-0.75, -0.75) rectangle (1.25, 1.25);
			\draw (0.25, -0.75) node[above] {\scriptsize $T_i(u^{-k})$};
			\draw[dashed] (0.75, 0.75) rectangle (2.75, 2.75);
			\draw (1.75, 2.75) node[above] {\scriptsize $T_i(u^{k})$};
			\draw[->, thick, gray] (-0.75, 1.25) -- (0.65, 2.65);
			\draw[->, thick, gray] (1.25, -0.75) -- (2.65, 0.65);
			\draw[->, thick, gray] (1.25, 1.25) -- (2.65, 2.65);
			\draw[->, thick, gray] (-0.75, -0.75) -- (0.65, 0.65);
			
			\node[inner sep=0pt] at (1, -3.5) {\includegraphics[width=4cm]{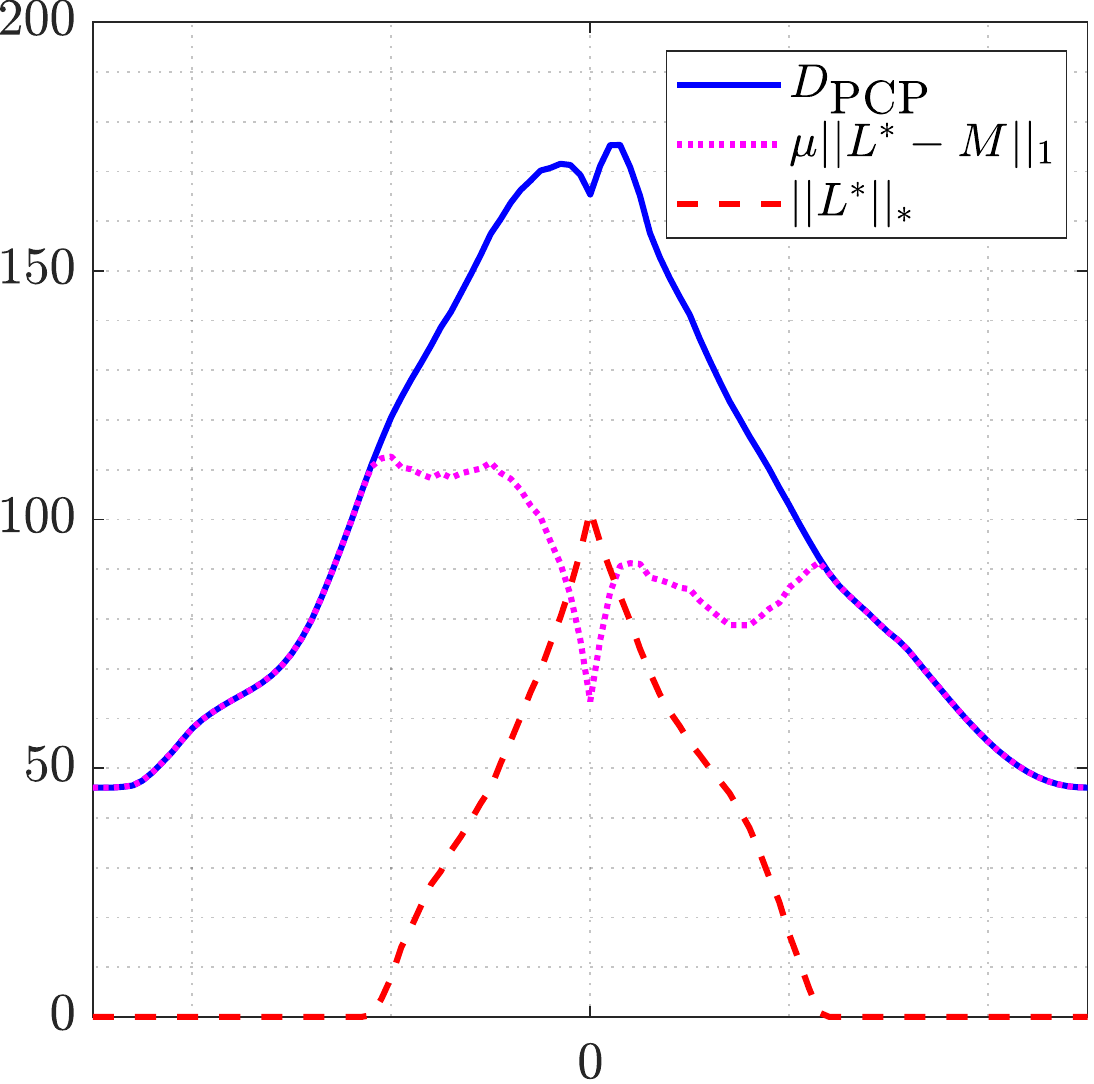}};
			\node[inner sep=0pt] at (1, -8) {\includegraphics[width=4cm]{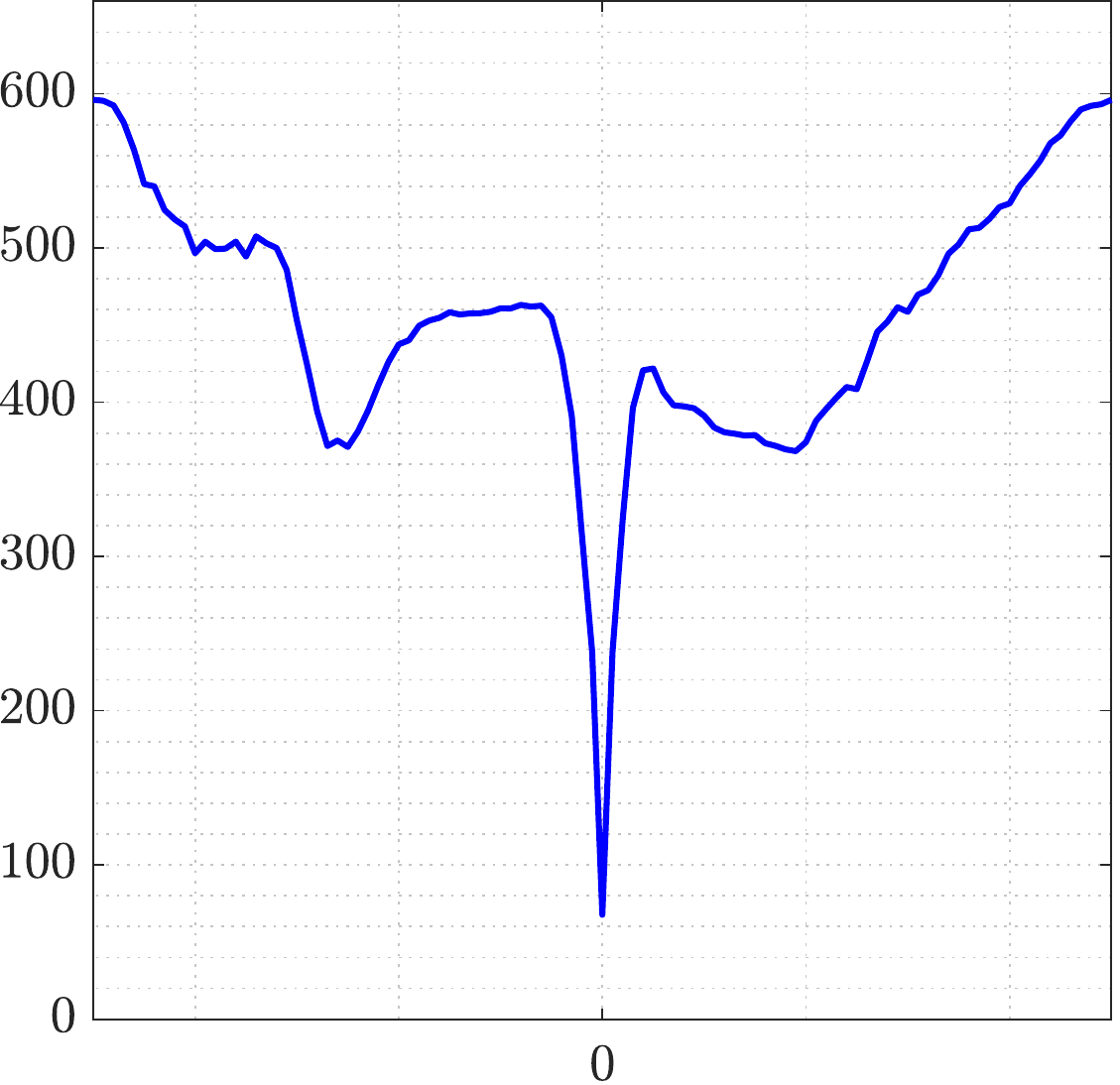}};
			
			\draw[ultra thick] (5, 0) rectangle (7, 2);
			\draw (6.95, 1.95) node[above right] {\scriptsize $T_1$};
			\draw[dashed] (6.5977, -0.2817) -- (4.7183, 0.4023) -- (5.4023, 2.2817) -- (7.2817, 1.5977) node[right] {\scriptsize $T_i(u^{-k})$} -- cycle;
			\draw[dashed] (5.6340, -0.3660) -- (4.634, 1.3660) -- (6.3660, 2.3660) node[above] {\scriptsize $T_i(u^k)$} -- (7.3660, 0.6340) -- cycle;
			\draw[->, thick, gray] (6,1) +(25:1.4142cm) arc [radius = 1.4142cm, start angle=25, delta angle=45];
			\draw[->, thick, gray] (6,1) +(115:1.4142cm) arc [radius = 1.4142cm, start angle=115, delta angle=45];
			\draw[->, thick, gray] (6,1) +(205:1.4142cm) arc [radius = 1.4142cm, start angle=205, delta angle=45];
			\draw[->, thick, gray] (6,1) +(295:1.4142cm) arc [radius = 1.4142cm, start angle=295, delta angle=45];
			
			\node[inner sep=0pt] at (6, -3.5) {\includegraphics[width=4cm]{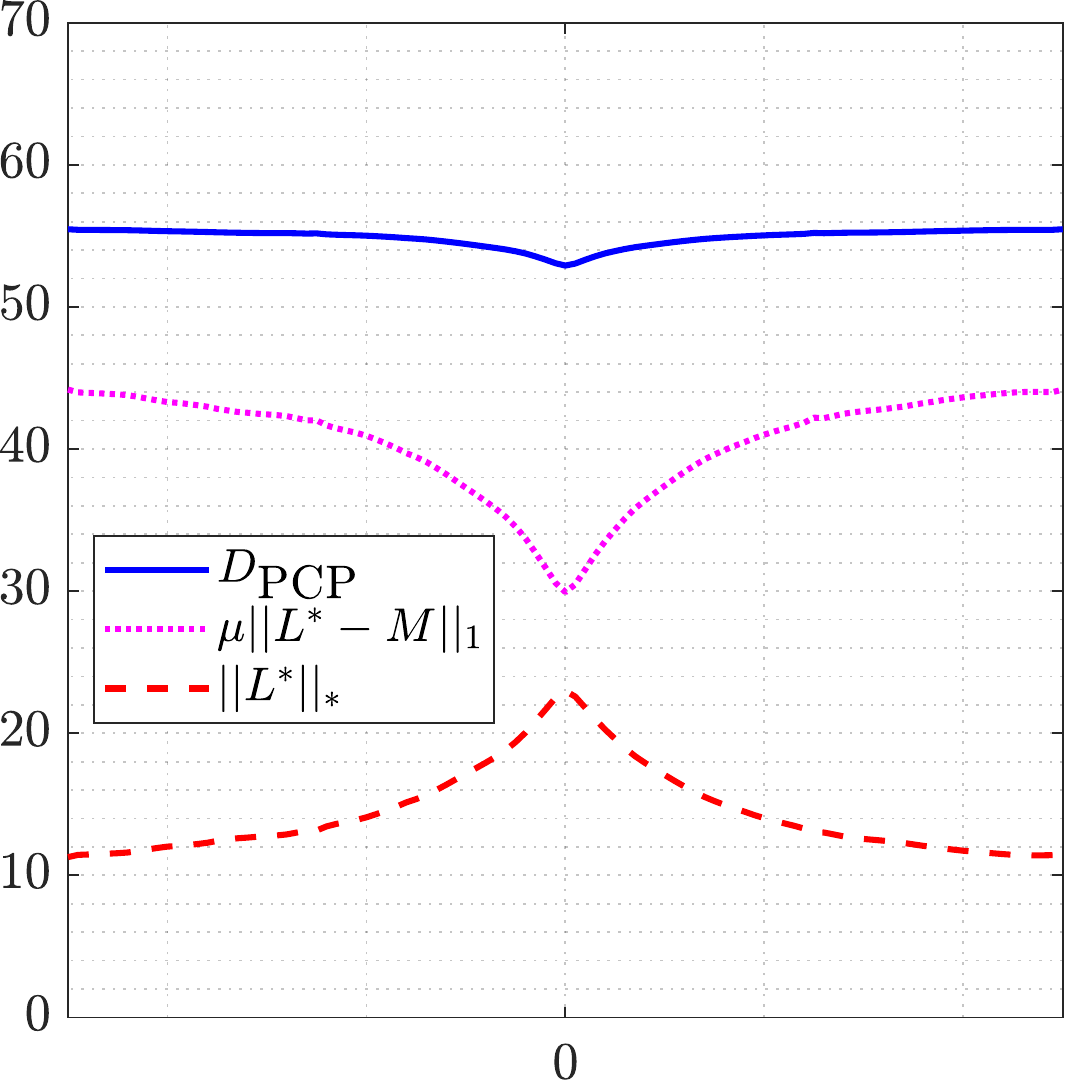}};
			\node[inner sep=0pt] at (6, -8) {\includegraphics[width=4cm]{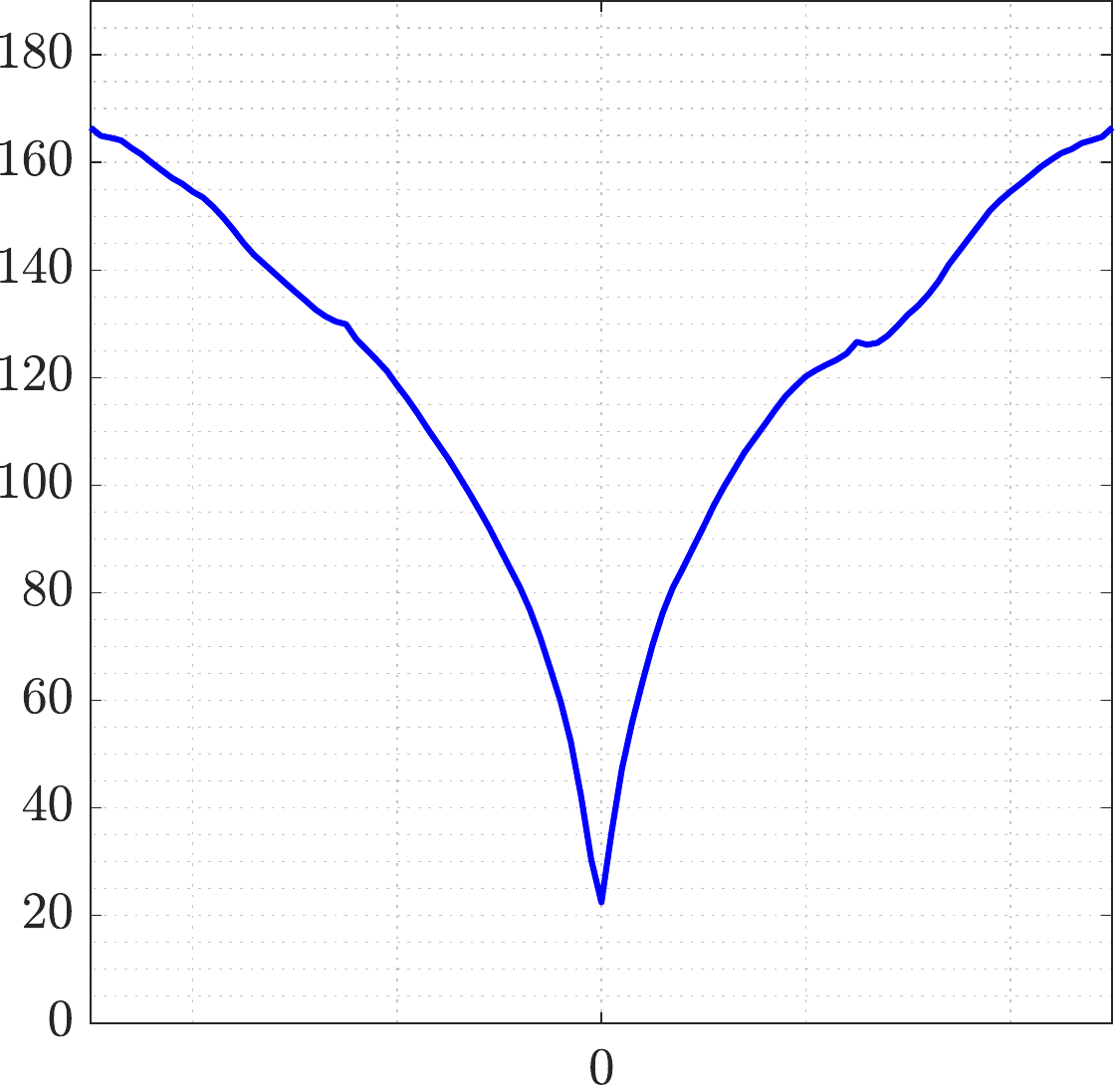}};
			
			\draw[ultra thick] (10, 0) rectangle (12, 2);
			\draw (11, 2) node[above] {\scriptsize $T_1$};
			\draw (11, 2.5) node[above] {\scriptsize $T_i(u^k)$};
			\draw (11, 1.5) node[above] {\scriptsize $T_i(u^{-k})$};
			\draw[dashed] (9.5, -0.5) rectangle (12.5, 2.5);
			\draw[dashed] (10.5, 0.5) rectangle (11.5, 1.5);
			\draw[<-, thick, gray] (9.6, -0.4) -- (10.5, 0.5);
			\draw[<-, thick, gray] (12.4, 2.4) -- (11.5, 1.5);
			\draw[<-, thick, gray] (12.4, -0.4) -- (11.5, 0.5);
			\draw[<-, thick, gray] (9.6, 2.4) -- (10.5, 1.5);
			
			\node[inner sep=0pt] at (11, -3.5) {\includegraphics[width=4cm]{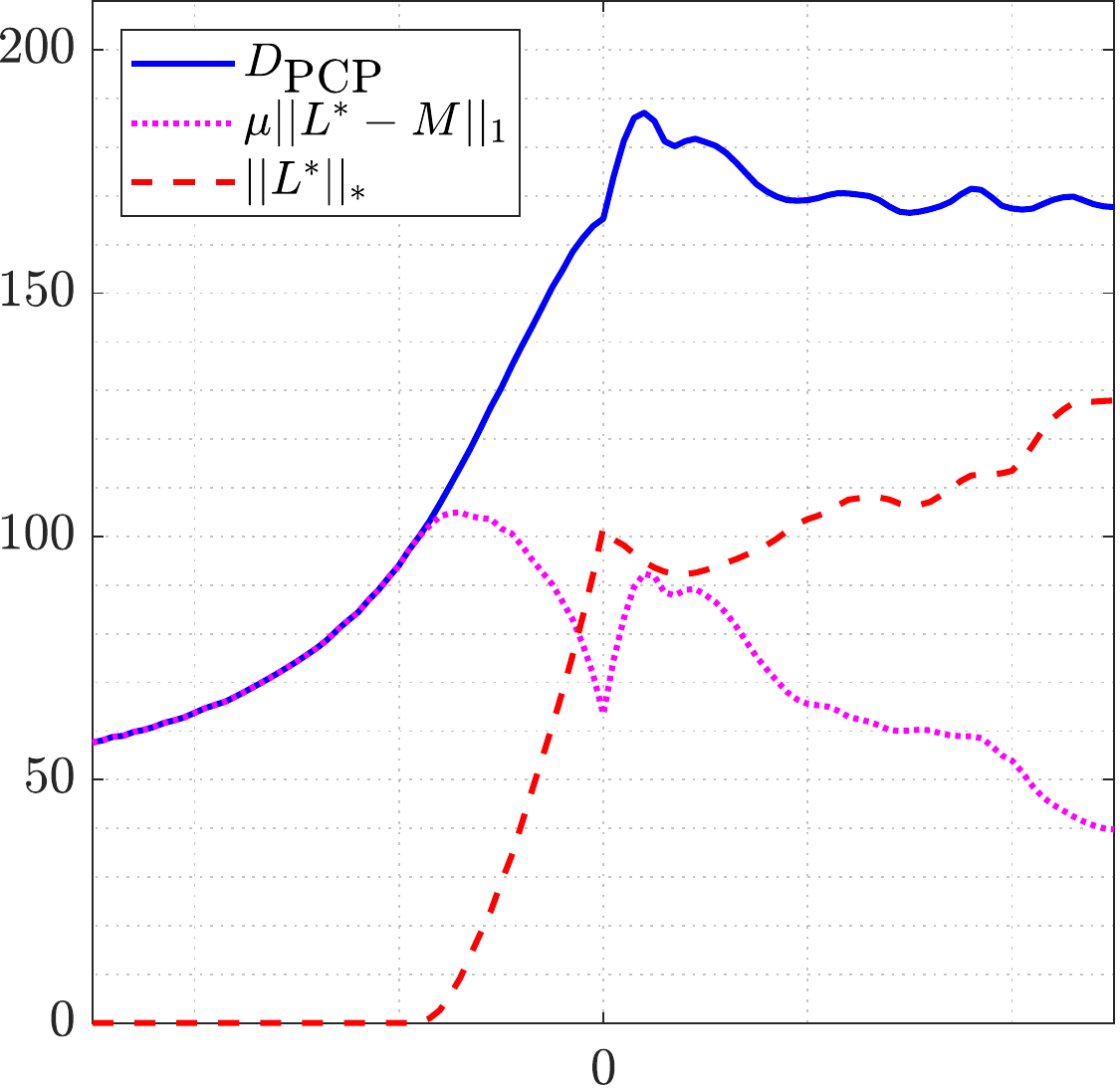}};
			\node[inner sep=0pt] at (11, -8) {\includegraphics[width=4cm]{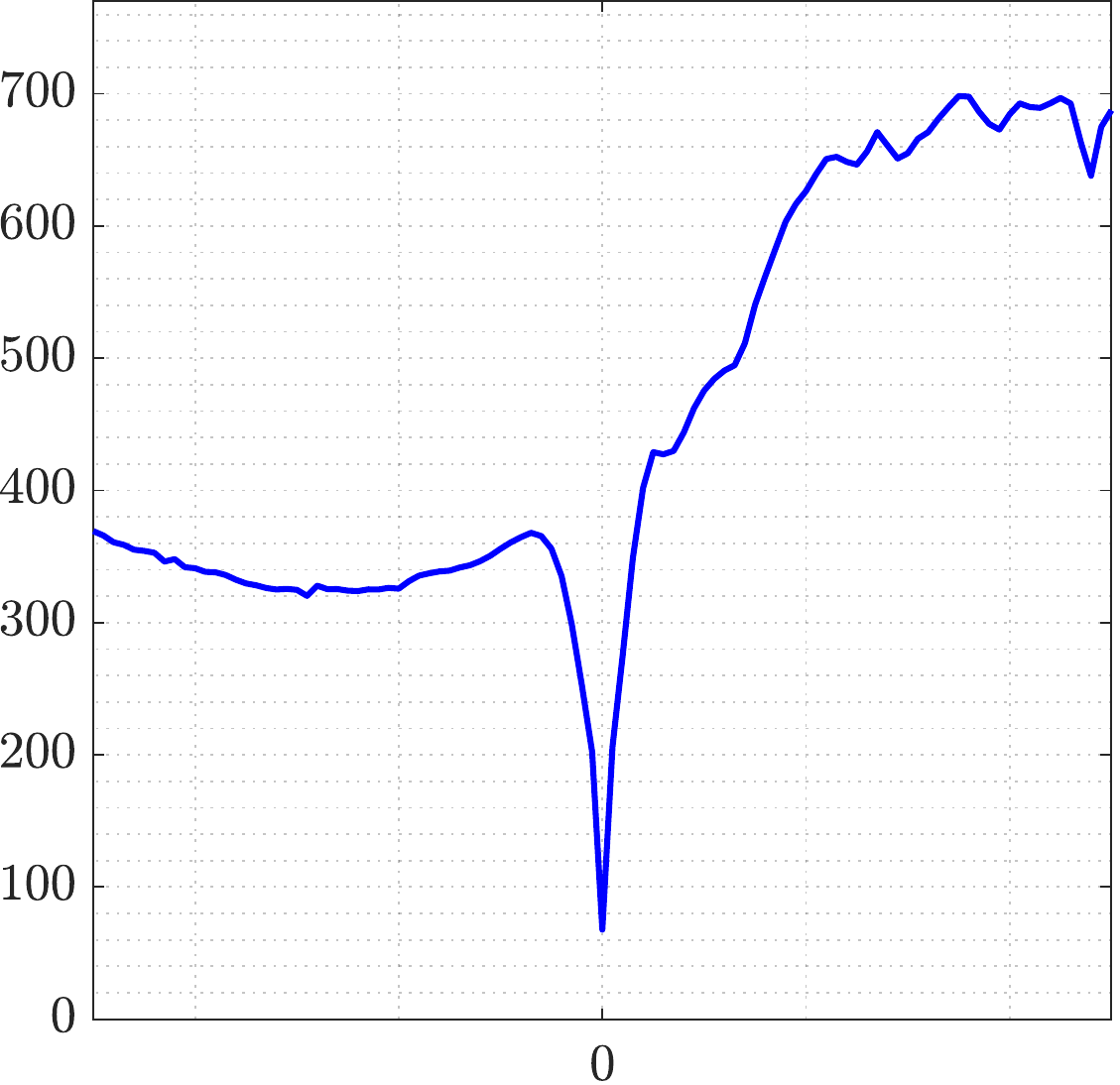}};
			
			\draw[ultra thick] (15, 0) rectangle (17, 2);
			\draw (16, 2) node[above] {\scriptsize $T_1$};
			\draw[dashed] (14.25, 0) -- (16.25, 0) -- (17.75, 2) node[above] {\scriptsize $T_i(u^{k})$} -- (15.75, 2) -- cycle;
			\draw[dashed] (15.75, 0) -- (17.75, 0) -- (16.25, 2) -- (14.25, 2) node[above] {\scriptsize $T_i(u^{-k})$} -- cycle;
			\draw[->, thick, gray] (16.475, 1.7) -- (17.425, 1.7);
			\draw[->, thick, gray] (14.475, 1.7) -- (15.425, 1.7);
			\draw[->, thick, gray] (17.525, 0.3) -- (16.575, 0.3);
			\draw[<-, thick, gray] (14.575, 0.3) -- (15.525, 0.3);
			
			\node[inner sep=0pt] at (16, -3.5) {\includegraphics[width=4cm]{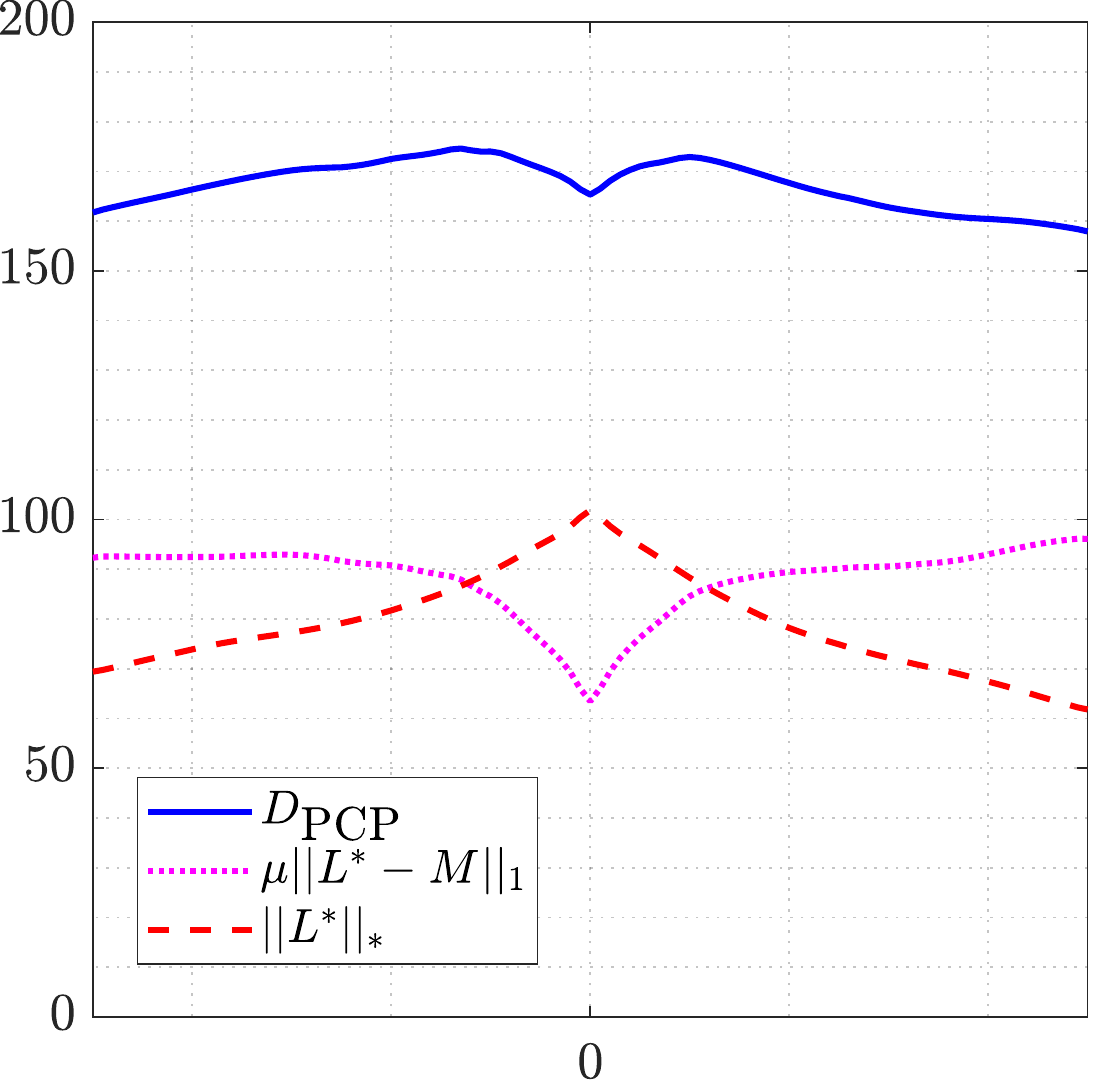}};
			\node[inner sep=0pt] at (16, -8) {\includegraphics[width=4cm]{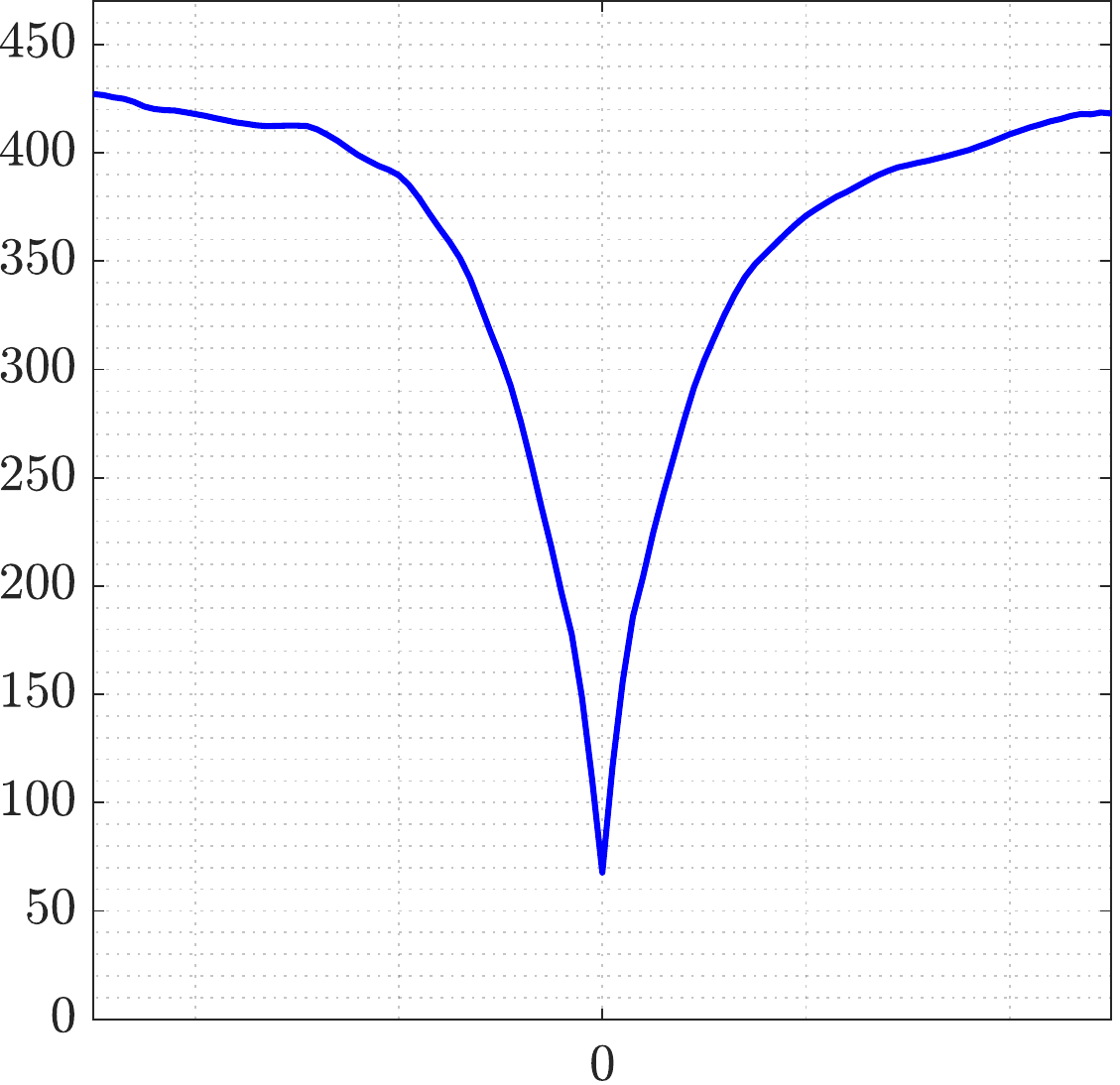}};

			\end{tikzpicture}
		}
		\caption{Experiments on the distance measures $D_{\text{PCP}}$ and $D_{\dRPCA}$.
			The classical $D_{\text{PCP}}$ energies \textbf{(second row)} exhibit only very narrow minima at the position $u^0 = 0$ of perfect alignment.
			Even worse so, \textit{global minimizers} for all experiments except the rotation are given by degenerated transformations.
			The proposed modification $D_{\dRPCA}$ \textbf{(third row)} resolves these problems: $u^0 = 0$ constitutes a global minimizer across all experiments.
			Furthermore, degenerated deformations generally result in high energies and are therefore not favored by this metric
		}
		\label{fig:distance_measure_exp}
	\end{figure}
	
	In order to assess the general applicability of \eqref{eq:pcp_measure} in the context of non-parametric groupwise registration, we conducted a number of experiments to examine the behavior of $D_{\text{PCP}}$ under certain predefined deformations of the images $T_1, \ldots, T_N$.
	
	To this end, we define an image $T \in \R^{m \times n}$ \textit{as a function of a deformation} $u \in \R^{m \times n \times 2}$ (given over the same grid) through linear interpolation -- see, e.g., \cite[Chapter 3.3]{Modersitzki2009}.
	Using this convention, the experiments were performed by evaluating $D_{\text{PCP}}(T_1, T_2(u^{j}), \ldots, T_N(u^{j}))$ for the four cases of the deformation sequence $(u^{j})_{j=-k, \ldots, k}$ describing a translation, a rotation, a scaling and a shearing.
	$T_1$ therefore acted as a fixed reference, while $T_2, \ldots, T_N$ were warped uniformly by $u^j$.
	
	A schematic depiction of each transformation sequence is given in the first row of Fig.~\ref{fig:distance_measure_exp}.
	Test data was comprised of the $N = 5$ frames from a cardiac MRI sequence, that are displayed in Fig.~\ref{fig:distance_measure_exp_frames}.
	As suggested by \cite[Theorem 1.1]{Candes2011}, the weighting parameter for \eqref{eq:pcp_measure} was chosen as $\mu = (m n)^{-1/2}$.
	
	The energy plots of $D_{\text{PCP}}$ for all four experiments are shown in the second row of Fig.~\ref{fig:distance_measure_exp}.
	Additionally, their respective decompositions into the two summands $||L^*||_*$ and $\mu ||M - L^*||_1$ are displayed, where $L^*$ denotes the minimizer of \eqref{eq:pcp_measure} over the variable $L$.
	
	The results show that $D_{\text{PCP}}$ has two major shortcomings that are unfavorable for the purposes of image registration.
	Firstly, the point $u^0 = 0$ at which the $N$ images are most appropriately aligned only marks a local minimizer of $D_{\text{PCP}}$ in a very narrow neighborhood of $u^0$ in the translation experiment.
	In the scaling experiment, $u^0$ even constitutes a \textit{global maximizer}.
	Secondly, both the left or the right endpoints of each energy plot, i.e., the most degenerated of all evaluated deformations $u^j$, represent global minimizers in every experiment except for the rotation.
	This is especially problematic in case of the translation, since constant translations are also not penalized by any regularizer, that is based on derivatives of deformations.
	As a result, we consider $D_{\text{PCP}}$ unsuitable as a distance metric in the general case.
	
	\subsection{Proposed Modified Measure}
	\label{subsec:modified_distance}
	
	Based on these observations, we propose a new distance metric that modifies $D_{\text{PCP}}$ in two aspects.
	As a first step, the $||\cdot||_*$-penalty term in \eqref{eq:pcp_measure} is replaced by a hard constraint to the set $\{|| \cdot ||_* \leq \nu\}$ for some suitable threshold $\nu \geq 0$.
	As a second step, we propose not to constrain the nuclear norm of $L$ itself, but to constrain that of the centered variable $L - \bar{L}$ instead.
	To this end, let $\bar{L} := (\sum_{i = 1}^{N} \frac{l_i}{N}) \cdot \mathbf{1}_{1 \times N} \in \R^{m n \times N}$ denote the matrix, in which every column is given by the average of the columns $l_i$ of $L$.
	
	The proposed dissimilarity measure, which we term $\delta$-RPCA, is then given by
	\begin{equation}
	D_{\dRPCA}(T_1, \ldots, T_N) :=
	\min_{L \in \R^{m n \times N}} ||M - L||_1 + \delta_{\{||\cdot||_* \leq \nu\}}(L - \bar{L}).
	\label{eq:mod_measure}
	\end{equation}
	
	Following the general convention in convex analysis \cite{Rockafellar1970,Rockafellar1998}, $\delta_{S}$ denotes an indicator function for a constraint set $S$, which is defined as $\delta_{S}(x) = 0$ for $x \in S$ and $\delta_{S}(x) = +\infty$ otherwise.
	
	The first modification is based on the observation that all energy curves of the $||\cdot||_*$-term in Fig.~\ref{fig:distance_measure_exp} exhibit at least a local maximum at $u^0$ and therefore counteract the local minimum of the $||\cdot||_1$-term in the joint $D_{\text{PCP}}$-energy.
	Remodeling the low-rank requirement on $L$ as a hard constraint resolves this issue by removing the nuclear norm from the energy as a summand.
	The second modification of centering $L$ further acts to model the low-rank requirement appropriately:
	
	Consider the nuclear norm of the two matrices $A_1 = a \cdot (1, 0, \ldots, 0) \in \R^{p \times q}$ and $A_2 = a \cdot \mathbf{1}_{1 \times q} \in \R^{p \times q}$ for some $a \in \R^p \setminus \{0\}$.
	While both matrices are obviously of rank one, a short derivation shows that one has
	\begin{equation}
	||A_1||_* = ||a||_2 < \sqrt{q} ||a||_2 = ||A_2||_*
	\label{eq:nn_relation_uncentered}
	\end{equation}
	for all $q > 1$.
	In terms of the registration model, this means that a smaller nuclear norm for the uncentered variable $L$ can be achieved by shifting all deformable images out of the image domain -- thereby replacing them with the boundary value of zero -- than by aligning them inside the images domain.
	
	The continuation of the above example shows that this situation, which is highly undesirable for the purpose of image registration, is reversed when dealing with centered variables.
	These are given by $A_1 - \bar{A_1} = a \cdot (1-q^{-1}, -q^{-1}, \ldots, -q^{-1})$ and $A_2 - \bar{A_2} = 0$ respectively and in fact, the equivalent of relation \eqref{eq:nn_relation_uncentered} now reads
	\begin{equation}
	||A_2 - \bar{A_2}||_*  = 0 < \sqrt{1 - q^{-1}} ||a||_2 = ||A_1 - \bar{A_1}||_*.
	\label{eq:nn_relation_centered}
	\end{equation}
	As a consequence, \eqref{eq:mod_measure} does not favor shifting the deformable images out of the image domain and is therefore more suited for image registration.
	
	Crucially, $D_{\dRPCA}$ is still convex in the variable $L$, since $\{|| \cdot ||_* \leq \nu\}$ constitutes the level set of a convex function and is therefore convex \cite[Theorem 4.6]{Rockafellar1970}.
	
	Repeating the above experiments for $D_{\dRPCA}$ with the choice of $\nu = 0.9 ||M - \bar{M}||_*$ for every set of deformed images in $M$, one obtains the energy plots in the third row of Fig.~\ref{fig:distance_measure_exp}.
	In contrast to $D_{\text{PCP}}$, the modified metric $D_{\dRPCA}$ shows global minimizers at the point $u^0 = 0$ across all cases.
	Additionally, the global maximizer of each curve is found towards its left or right endpoint and consequently results from a degenerated deformation.
	In conclusion, the two main issues of $D_{\text{PCP}}$ as a distance function for registration tasks are hence resolved by $D_{\dRPCA}$.
	
	Apart from the interpretation of $D_{\dRPCA}$ as a modified version of $D_{\text{PCP}}$, it can also be interpreted in the sense described as follows.
	First consider the case $\nu = 0$.
	This implies $L = \bar{L}$, which in turn implies constant columns $l_1 = \ldots = l_N$ of $L$.
	Using this, one can solve \eqref{eq:mod_measure} analytically for $L$ by recalling, that $\ell_1$-distance minimization problems of the type
	\begin{equation}
	\argmin_{x \in \R} \sum_{i = 1}^K |x - y_i|
	\end{equation}
	are solved by the median of $(y_1, \ldots, y_K)$ \cite[p. 433]{Boyd2004}.
	As a consequence, the constant columns of $L$ in the problem above are given by the pointwise median of $T_1, \ldots, T_N$ and $D_{\dRPCA}$ represents the remaining $\ell_1$-distance between the input images and that median.
	
	In the case of $\nu > 0$, $D_{\dRPCA}$ can now more generally be interpreted as the joint $\ell_1$-distance between the images $T_1, \ldots, T_N$ and their individual (optimal) approximations $l_1, \ldots, l_N$ with deviations from the mean $\bar{l} := \sum_{i = 1}^N \frac{l_i}{N}$ restricted to a low-dimensional linear subspace.
	
	Consequently, we deem \eqref{eq:mod_measure} especially suited for image groups with inherent low-dimensional structure such as image sequences with strong or pronounced temporal repetition.
	
	\section{Regularization}
	\label{sec:regularization}
	
	\subsection{Total Variation Regularization}
	\label{subsec:tv}
	Total variation (TV) is a popular choice for regularizing motion fields in applications of both optical flow estimation and image registration due to its distinguishing feature of allowing discontinuities in the solution.
	TV therefore sets itself apart from other common regularizers such as diffusive, elastic or curvature energies that favor smooth transformations.
	Exemplary early applications of TV regularization for optical flow estimation can be found in \cite{Papenberg2006,Zach2007} and for image registration in \cite{Pock2007,Vishnevskiy2017}.
	In the context of medical image processing, TV regularization is particularly interesting when modeling non-smooth sliding motions, since it eliminates the necessity to explicitly mask all sliding interfaces beforehand \cite{Derksen2015}.
	
	We shortly recapitulate that the total variation for vector fields $\upsilon \in L^1(\Omega, \R^d)$ over $\Omega \subset \R^d$ can be defined as
	\begin{equation}
	\TV(\upsilon) = \int_{\Omega} \diff |D \upsilon|,
	\label{eq:tv_distributional}
	\end{equation}
	where $D \upsilon$ denotes the distributional (measure-valued) derivative of $\upsilon$ with values in $\R^{d \times d}$.
	In case of $\upsilon \in C^1(\Omega, \R^d)$, \eqref{eq:tv_distributional} is equivalent to
	\begin{equation}
	\TV(\upsilon) = \int_{\Omega} ||\nabla \upsilon||_2 \diff x.
	\label{eq:tv_smooth}
	\end{equation}
	For all further details, we refer to \cite{Ambrosio2000}.
	
	In our registration model, we assume rectangular domains $\Omega \subset \R^2$ and employ a standard discretization scheme with cell-centered grids of resolution $m \times n$ and grid spacings of $(h_1, h_2) \in \R_{> 0}^2$ in the two coordinate directions.
	Optimization is performed over discrete displacement fields $u^k \in \R^{m \times n \times 2}$, for which we use finite forward differences and Neumann boundary conditions to discretize \eqref{eq:tv_smooth}.
	Following \cite{Vishnevskiy2017}, we use the notation
	\begin{equation}
	||v||_{2, 1} := \sum_{i = 1}^{p} || (v_i, v_{i + p}, v_{i + 2 p}, v_{i + 3 p}) ||_2
	\label{eq:norm_2_1}
	\end{equation}
	for $v \in \R^{4 p}$ and obtain a discretization of \eqref{eq:tv_smooth}
	\begin{equation}
	\TV^h(u^k) := h_1 h_2 || G \vect(u^k) ||_{2, 1}.
	\label{eq:tv_discrete}
	\end{equation}
	Therein, $G \in \R^{4 m n \times 2 m n}$ denotes the finite difference operator with the aforementioned characteristics.
	
	\subsection{Uniqueness Constraint}
	\label{subsec:uniqueness}
	As our model does not make use of an explicit reference image that all other images are aligned to, we need to employ an additional constraint on the displacements $u^1, \ldots, u^N$ in order to ensure the uniqueness of a solution.
	
	This can be seen from the simple example, in which $T_1, \ldots, T_N$ display uniform objects, e.g., white rectangles, before a black background.
	Consider the case of a perfect alignment $T_1(u^1) = \ldots = T_N(u^N)$ of these rectangles inside the image domain $\Omega$.
	If all deformations $u^k$ are simultaneously offset by $t \in \R^2$, such that the new deformations $\hat{u}^k$ still align $T_1, \ldots, T_N$ inside the common domain\footnote{To be exact, these are defined as $\hat{u}_{i,j,c}^k := u_{i,j,c}^k + t_c$ for all $i = 1, \ldots, m$, $j = 1, \ldots, n$, $c = 1, 2$ and $k = 1, \ldots, N$.}, then $(\hat{u}^k)_{k=1,\ldots,N}$ constitute a solution equal to $(u^k)_{k=1,\ldots,N}$ both in terms of $D_{\dRPCA}$ and $\TV^h$.
	
	For $\TV^h$, this is explained by \eqref{eq:tv_discrete} solely penalizing derivatives of deformation fields which are always invariant to translations.
	
	The invariance for $D_{\dRPCA}$ is due to the equivalence of an offset by $t$ and a simple reordering of the pixels between $T_k(u^k)$ and $T_k(\hat{u}^k)$ (due to the zero boundary condition).
	Clearly, the $\ell_1$-term in \eqref{eq:mod_measure} is invariant to any reordering and the same is true for the nuclear norm constraint, since a consistent reordering of all $T_k(u^k)$ results in a row permutation of the Casorati matrix $M = [\vect(T_1(u^1)) | \ldots | \vect(T_N(u^N))]$.
	As a short derivation shows, a row permutation does not affect the singular values of a matrix:
	
	Let $A \in \R^{p \times q}$ be an arbitrary matrix and let $P \in \{0, 1\}^{p \times p}$ be a permutation.
	If a singular value decomposition (SVD) of $A$ is given by $A = U \Sigma V^{\top} $, then $P A = (P U) \Sigma V^{\top}$ constitutes a valid SVD of $P A$ due to $P U$ still being orthogonal, i.e.,
	\begin{equation}
	(P U)^{\top} (P U) = U^{\top} P^{\top} P U = U^{\top} U = I.
	\label{eq:svd_permutation}
	\end{equation}
	Thus, the singular values on the diagonal of $\Sigma$ stay unaffected and so does the nuclear norm $||P A||_* = ||A||_*$.
	
	In order to eliminate this remaining degree of freedom from the model, we impose an additional constraint on the deformations $u^1, \ldots, u^N$, enforcing the mean (or equivalently the sum) over all deformations and grid points to be zero in each coordinate direction:
	\begin{equation}
	\frac{1}{(N m n)} \sum_{k = 1}^N \sum_{i = 1}^m \sum_{j = 1}^n u_{i, j, c}^k \overset{!}{=} 0 \quad \forall c \in \{1, 2\}.
	\label{eq:uniqueness_constraint}
	\end{equation}
	
	Note that \cite{Metz2011,Huizinga2016,Guyader2018,Polfliet2018} constrain their deformations in a related manner by demanding the mean of all deformations to be zero \textit{at every grid point} as first introduced by \cite{Bhatia2004}.
	The difference however is, that \eqref{eq:uniqueness_constraint} only imposes one constraint \textit{per dimension} instead of one constraint \textit{per grid point and dimension}.
	As a result, \eqref{eq:uniqueness_constraint} restricts the space of feasible solutions much less severely while still ensuring uniqueness.
	
	\section{Implementation \& Optimization}
	\label{sec:optimization}
	In this section, we present an optimization scheme for our groupwise registration model that is strongly related to the work in \cite{Heber2014}.
	First we combine all components derived in the previous sections into the complete registration model
	\begin{align}
	\begin{split}
	\min_{\substack{u^1, \ldots, u^N \in \R^{m \times n \times 2} \\ L \in \R^{m n \times N}}} \ &|| [\vect(T_1(u^1)), \ldots, \vect(T_N(u^N)] - L ||_1 + \delta_{\{||\cdot||_* \leq \nu\}}( L - \bar{L} ) \\[-4ex]
	& + \mu \sum_{k = 1}^N \TV^h(u^k) + \sum_{c = 1}^2 \delta_{ \{ \langle \mathbf{1}, \cdot \rangle = 0 \} } ((u_{\bigcdot, \bigcdot, c}^1, \ldots, u_{\bigcdot, \bigcdot, c}^N)),
	\label{eq:full_model}
	\end{split}
	\end{align}
	in which $\mu > 0$ controls the regularization strength.
	
	\subsection{Linearized Subproblems}
	\label{subsec:linearization}
	In order to be able to apply convex optimization methods to \eqref{eq:full_model}, one needs to deal with the non-linearity of the expressions $T_k(u^k)$ that leads to a non-convexity of the model.
	As in \cite{Pock2007,Zach2007,Peng2010,Heber2014}, an iterative linearization of the deformed images is used to overcome this issue.
	Note that while a one-time linear approximation would also be possible in theory, the strong locality of such an approximation becomes prohibiting when larger deformations are required to align the images.
	
	In the following, we assume all variables to be in vector format (including the values of all $T_k$) and for brevity's sake omit the explicit notation of reshaping operations like $\vect(\cdot)$.
	The linearization of $T_k$ can then be expressed as
	\begin{equation}
	T_k(u^k) \approx T_k(\tilde{u}^k) + \nabla T_k(\tilde{u}^k)^{\top} \cdot (u^k - \tilde{u}^k)
	\end{equation}
	for a suitable point $\tilde{u}^k$.
	This enables one to approximate the first term in \eqref{eq:full_model} by
	\begin{equation}
	\sum_{k = 1}^N || T_k(\tilde{u}^k) + \nabla T_k(\tilde{u}^k)^{\top} \cdot (u^k - \tilde{u}^k) - l_k ||_1.
	\end{equation}
	Using vectorized variables further allows one to rewrite the centering of $L$ as a linear operation $K L$ with
	\begin{equation}
	K = \left( I_{N \times N} - \frac{\mathbf{1}_{N \times N}}{N} \right) \otimes I_{m n \times m n} \in \R^{m n N \times m n N}.
	\label{eq:center_operator}
	\end{equation}
	
	Since solving \eqref{eq:full_model} through iterative (re-)linearization amounts to solving a series of subproblems, we propose to treat is as a process, in which the threshold $\nu$ is successively decreased to the threshold value for which the original problem \eqref{eq:full_model} is meant to be solved.
	
	Assuming a predefined number $n_{iter}$ of linearization steps and denoting the final threshold by $\nu$, we therefore employ a series of thresholds $\nu_1 > \nu_2 > \ldots > \nu_{n_{iter}} = \nu$ for the iterative solution of the separate subproblems.
	As a strategy to select these parameters, we propose to choose $\nu$ relative to the nuclear norm of the centered input images and to progressively decrease $\nu_t$ to that value by multiplication with a constant factor $\alpha \in (0, 1)$.
	
	More specifically, let $M$ denote the Casorati matrix of the input images and let $\bar{M}$ denote the columnwise repetition of their mean.
	If one now wants to meet a final threshold of $\nu = \beta || M - \bar{M} ||_*$ for some $\beta \in (0, 1)$ and one employs a predefined number of $n_{iter}$ linearization steps, the proposed strategy amounts to choosing
	\begin{equation}
	\nu_t = \alpha^t || M - \bar{M} ||_* \quad \mbox{for } t = 1, \ldots, n_{iter},
	\end{equation}
	where $\alpha = \beta^{(1 / n_{iter})}$.
	
	\subsection{Solving the Convex Subproblem}
	\label{subsec:cvx_subproblem}
	Denoting all entries of $u^k$ corresponding to the $c$-th coordinate axis by $u^{k, c}$ ($u_{\bigcdot, \bigcdot, c}^k$ in the non-vectorized notation of \eqref{eq:full_model}), the $t$-th subproblem now reads
	\begin{align}
	\begin{split}
	\min_{\substack{u^1, \ldots, u^N \\ L }} & \sum_{k = 1}^N || T_k(\tilde{u}^k) + \nabla T_k(\tilde{u}^k)^{\top} \cdot (u^k - \tilde{u}^k) - l_k ||_1 + \delta_{\{||\cdot||_* \leq \nu_t\}}( K L )\\[-2ex]
	& + \mu \sum_{k = 1}^N h_1 h_2 || G u^k ||_{2, 1} + \sum_{c = 1}^2 \delta_{ \{ \langle \mathbf{1}, \cdot \rangle = 0 \} } ((u^{1, c}, \ldots, u^{N, c})).
	\label{eq:subproblem}
	\end{split}
	\end{align}
	We solve these subproblems using the primal-dual optimization algorithm~\ref{alg:cp} from \cite{Chambolle2011} that is designed for finding saddle-points of problems of the type
	\begin{equation}
	\min_{x \in \R^p} \max_{y \in \R^q} \ \langle A x, y \rangle + H(x) - F^*(y).
	\label{eq:saddle_point_problem}
	\end{equation}
	$H: \R^p \rightarrow (\R \cup \{ \infty \} =: \bar{\R})$, $F^*: \R^q \rightarrow \bar{\R}$ are proper, lower-semi\-continuous, convex functions and $F^*$ denotes the conjugate of another proper, lower-semi\-continuous, convex function $F: \R^q \rightarrow \bar{\R}$.
	$A \in \R^{q \times p}$ further denotes a linear operator.
	As is well-known, \eqref{eq:saddle_point_problem} is equivalent to the primal minimization problem
	\begin{equation}
	\min_{x \in \R^p} \ F(A x) + H(x).
	\label{eq:primal_problem}
	\end{equation}
	For all details, we refer to \cite[Chapter 11]{Rockafellar1998}.
	
	\begin{algorithm}[t]
		\SetKw{Initialization}{Initialization}
		\SetKwFor{While}{Iterate over}{:}{ }
		\SetAlgorithmName{Alg.}{}
		
		\Initialization{\\Choose $x^0 \in \R^{p}$, $y^0 \in \R^{q}$. Set $\bar{x}^0 \leftarrow x^0$. Choose $\tau, \eta > 0$ s.t. $\tau \eta || A ||_{\sigma}^2 < 1$ where $|| A ||_{\sigma} := \max \{ || A x ||_2 : x \in \R^p, ||x||_2 \leq 1 \}$}\\
		\While{$n \geq 0$}{
			\vspace{-1em}
			\begin{flalign}
			&y^{n + 1} \leftarrow (\id + \eta \partial F^*)^{-1} (y^n + \eta A \bar{x}^n) & \nonumber \\
			&x^{n + 1} \leftarrow (\id + \tau \partial H)^{-1} (x^n - \tau A^{\top} y^{n+1}) & \label{eq:cp_updates} \\
			&\bar{x}^{n+1} \leftarrow 2 x^{n+1} - x^n & \nonumber
			\end{flalign}
			\vspace{-1.5em}
		}
		\caption{Primal-dual Optimization Scheme \cite{Chambolle2011}}
		\label{alg:cp}
	\end{algorithm}
	
	In our case, we bring \eqref{eq:subproblem} into the form \eqref{eq:primal_problem} by assigning the first three terms of \eqref{eq:subproblem} to $F$ and the remaining uniqueness term to $H$.
	The primal variables for this problem are given by the union of all variables over which \eqref{eq:subproblem} is minimized,
	\begin{equation}
	x^{\top} = \left[ (u^1)^{\top}, \ldots, (u^N)^{\top}, (l_1)^{\top}, \ldots, (l_N)^{\top} \right] \in \R^{3 N m n},
	\end{equation}
	and we define the linear operator $A$ to be
	\begin{equation}
	\mbox{\LARGE $A := $} \left[
	\begin{array}{c c c c c c}
	\nabla T_1(\tilde{u}^1)^{\top} &  &  & -I_{m n} &  & \\
	& \ddots &  &  & \ddots & \\
	&  & \nabla T_N(\tilde{u}^N)^{\top} &  &  & -I_{m n} \\[.5em]
	
	\multicolumn{3}{c}{ \mbox{\LARGE $0$} } & \multicolumn{3}{c}{ \mbox{\LARGE $K$} } \\[.5em]
	
	G &  &  & \multicolumn{3}{c}{\multirow{3}{*}{ \mbox{\LARGE $0$} }} \\
	& \ddots &  &  &  & \\
	&  & G &  &  & \\ 
	\end{array}
	\right],
	\label{eq:big_A}
	\end{equation}
	where $K$ and $G$ are as in \eqref{eq:center_operator} and \eqref{eq:tv_discrete} respectively.
	
	This allows for a separable definition of the function~$F$ by $F(z) := F_1(z^1) + F_2(z^2) + F_3(z^3)$ and
	\begin{align}
	F_1(z^1) &:= || z^1 + b ||_1, \label{eq:F_1}\\
	F_2(z^2) &:= \delta_{\{||\cdot||_* \leq \nu_t\}}(z^2), \label{eq:F_2}\\
	F_3(z^3) &:= \mu \sum_{k = 1}^N h_1 h_2 || z^{3, k} ||_{2, 1}, \label{eq:F_3}
	\end{align}
	where $z^{3, k} := (z^3_{4 (k - 1) m n + 1}, \ldots, z^3_{4 k m n})^{\top}$ and where
	the vector $b = (b_1^{\top}, \ldots, b_N^{\top})^{\top}$ gathers the constants $b_k := T_k(\tilde{u}^k) - \nabla T_k(\tilde{u}^k)^{\top} \tilde{u}^k$ for $k = 1, \ldots, N$.
	As the remaining uniqueness term does not depend on $l_1, \ldots, l_N$, we define $H(x) := \tilde{H}( (u^1, \ldots, u^N) )$ and
	\begin{equation}
	\tilde{H}( (u^1, \ldots, u^N) ) := \sum_{c = 1}^2 \delta_{ \{ \langle \mathbf{1}, \cdot \rangle = 0 \} } ((u^{1, c}, \ldots, u^{N, c})).
	\label{eq:H_1}
	\end{equation}
	
	In order to apply Alg.~\ref{alg:cp} to the problem, the proximal operators $(\id + \eta \partial F^*)^{-1}$ and $(\id + \tau \partial H)^{-1}$ are required to compute the updates \eqref{eq:cp_updates}.
	We point out that the separable nature of $F$ implies a decomposability of $(\id + \eta \partial F^*)^{-1}$ into three terms corresponding to the three summands of $F$ (or equivalently of $F^*$). 
	These terms as well as the proximal operator of $H$ are given by standard expressions, for which we refer to \cite{Parikh2014}.
	
	We however emphasize the point that the proximal step corresponding to the nuclear norm constraint
	\begin{equation}
	(\id + \eta \partial F_2^*)^{-1}(y) =\\
	U \diag \left( \sigma - \eta \nu_t \Pi_{ \{ ||\cdot||_1 \leq 1 \} } \left( \frac{\sigma}{\eta \nu_t} \right) \right) V^{\top}
	\label{eq:F_3_star_prox}
	\end{equation}
	requires both an SVD $y = U \diag(\sigma) V^{\top}$, $\sigma \in \R^N$, of the input $y$ (assumed to be $m n \times N$-shaped) and a projection $\Pi_{ \{ ||\cdot||_1 \leq 1 \} }$ onto the $\ell_1$-unit ball.
	While the latter step cannot be solved in a decoupled manner, there exist exact algorithms with time complexity $\mathcal{O}(N)$ to compute such projections -- in our implementation we employ the approach from \cite{Duchi2008}.
	
	Finally, we shortly address the problem of determining the spectral norm $||A||_{\sigma}$ that the primal and dual step sizes $\tau, \eta$ for Alg.~\ref{alg:cp} are based on.
	Since the linear operator $A$ given by \eqref{eq:big_A} contains the image gradients $\nabla T_k(\tilde{u}^k)$ and is therefore dependent on empirical data, an analytical solution for $||A||_{\sigma}$ is unattainable.
	Instead, we apply a simple \textit{power iteration scheme} to estimate this quantity \cite[Section 7.3.1]{Golub1996}.
	
	\subsection{Multilevel Scheme and Parameter Scaling}
	\label{subsec:multilevel}
	
	\begin{algorithm*}[t]
		
		\SetKw{Initialization}{Initialization}
		\SetKw{Update}{Update}
		\SetKw{Estimate}{Estimate}
		\SetKw{Solve}{Solve}
		\SetKw{Prolongate}{Prolongate}
		\SetKwFor{For}{For}{:}{ }
		\SetKwIF{If}{ElseIf}{Else}{If}{:}{ }{ }{ }
		\SetAlgorithmName{Alg.}{}
		
		\Initialization{\\$\tilde{x}, \tilde{y} \leftarrow 0$; $\nu \leftarrow 2^{-n_{lev}} || [T_1 | \ldots | T_N ] - (\sum_{k = 1}^N \frac{T_k}{N}) \cdot \mathbf{1}_{1 \times N} ||_*$\\
			Choose $\alpha, \mu > 0$.}
		
		\vspace{.5em}
		\tcc{Outer Iteration: Problem Scaling + Prolongation}
		\For{$j = 1, \ldots, n_{lev}$}{
			\Update grid widths $h_1, h_2 \leftarrow 2^{(n_{lev} - j)}$.\\ 
			\Update threshold scale $\nu \leftarrow 2 \nu$. 
			
			\vspace{.5em}
			\tcc{Inner Iteration: (Re-)Linearization Process + Solving Convex Subproblems\\ \qquad for $(n_{lev} -j)$-fold downsampled images $T_1, \ldots, T_N$}
			\For{$k = 1, \ldots, n_{iter}^j$}{
				\Update threshold $\nu \leftarrow \alpha \nu$.\\ 
				\Estimate $||A||_2$ $\rightarrow$ choose $\tau, \eta$ s.t. $\tau \eta ||A||_{\sigma}^2 < 1$.\\
				\Solve
				$ \min_{x} \max_{y} \ \langle A x, y \rangle + H(x) - F^*(y) $
				for $x^*, y^*$ using Alg.~\ref{alg:cp} with starting points $\tilde{x}, \tilde{y}$.\\
				\Update starting points $\tilde{x} \leftarrow x^*$, $\tilde{y} \leftarrow y^*$.\\
				\Update linearization points $\tilde{u}^1, \ldots, \tilde{u}^N$ from $x^*$. 
			}
			\vspace{.5em}
			\If{$j < n_{lev}$}{
				\Prolongate $\tilde{x}, \tilde{y}$ as in Fig.~\ref{fig:prolongation}.
			}
		}
		
		\caption{Multilevel Scheme}
		\label{alg:ml}
	\end{algorithm*}
	
\begin{figure}[b]
		\centering
		\resizebox{.6\columnwidth}{!}{
			\begin{tikzpicture}
			
			\draw[step = 1, gray!50!white, thin] (0.75, 0.75) grid (4.25, 4.25);
			\filldraw[lightgray] (2.05, 2.05) rectangle (2.95, 2.95);
			
			\draw[step = 0.5, gray!50!white, thin] (5.75, 0.75) grid (9.25, 4.25);
			\filldraw[lightgray] (7.05, 2.05) rectangle (7.45, 2.45);
			\filldraw[lightgray] (7.55, 2.05) rectangle (7.95, 2.45);
			\filldraw[lightgray] (7.05, 2.55) rectangle (7.45, 2.95);
			\filldraw[lightgray] (7.55, 2.55) rectangle (7.95, 2.95);
			
			\node (1) at (2.5, 2.5) {};
			\node (2) at (7.25, 2.25) {};
			\node (3) at (7.75, 2.25) {};
			\node (4) at (7.25, 2.75) {};
			\node (5) at (7.75, 2.75) {};
			
			\path (1) edge[bend right, ->] (2);
			\path (1) edge[bend right, ->] (3);
			\path (1) edge[bend left, ->] (4);
			\path (1) edge[bend left, ->] (5);
			
			\draw (7.25, 2.25) node[above left=-4pt] {\scriptsize $(2 i, 2 j\!-\!1)\,$};
			\filldraw (7.25, 2.25) circle [radius = 0.8pt];
			
			\draw (7.75, 2.25) node[right] {\scriptsize $(2 i, 2 j)$};
			\filldraw (7.75, 2.25) circle [radius = 0.8pt];
			
			\draw (7.75, 2.75) node[right] {\scriptsize $(2 i\!-\!1, 2 j)$};
			\filldraw (7.75, 2.75) circle [radius = 0.8pt];
			
			\draw (7.25, 2.75) node[below left=-4pt] {\scriptsize $(2 i\!-\!1, 2 j\!-\!1)\,$};
			\filldraw (7.25, 2.75) circle [radius = 0.8pt];
			
			\draw (2.5, 2.5) node[above left=-3pt] {\scriptsize $(i, j)$};
			\filldraw (2.5, 2.5) circle [radius = 0.8pt];
			
			\end{tikzpicture}
		}
		
		\caption{Prolongation scheme. Variable values for the index $(i, j)$ in the low-resolution coordinate system \textbf{(left)} are propagated to the variables indexed by ${(2 i - 1, 2 j - 1)}$, ${(2 i - 1, 2 j)}$, ${(2 i, 2 j - 1)}$, ${(2 i, 2 j)}$ in the high-resolution system \textbf{(right)}}
		\label{fig:prolongation}
	\end{figure}
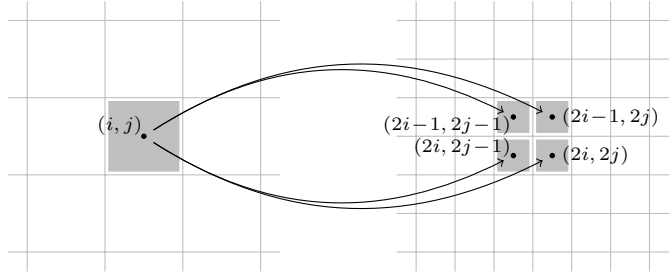

	As is common in image registration, we couple the techniques discussed in the previous subsections \ref{subsec:linearization} and \ref{subsec:cvx_subproblem} with a \textit{multilevel scheme}.
	This serves the two purposes of lowering the computational effort of our solution strategy on the one hand and of avoiding local minimizers on the other hand \cite{Modersitzki2009}.
	An image pyramid of $n_{lev}$ resolution stages serves as input to our multilevel scheme, where images are downsampled by a factor of $2$ in each dimension between consecutive stages (for ease of presentation we assume $2^{(n_{lev} - 1)} \mid m$, $2^{(n_{lev} - 1)} \mid n$).
	The inverse operation, i.e., the prolongation of a variable, is implemented as depicted in Fig.~\ref{fig:prolongation}.
	
	In order to guarantee a consistent scaling of all parts of the subproblem energy \eqref{eq:subproblem} between the different resolutions, we introduce an additional scaling of the $\ell_1$-term from \eqref{eq:F_1}, i.e., we redefine it as
	\begin{equation}
	F_1(z^1) := h_1 h_2 || z^1 + b ||_1.
	\end{equation}
	
	Moreover, a consistent scaling of the thresholds $\nu_t$ is required since the low-rank components change in resolution as well in between stages.
	To this end, it is useful to derive by what factor the nuclear norm of a matrix $M \in \R^{p \times q}$ scales when it is prolongated to the next higher resolution.
	Interpreting the columns of $M$ as vectorized images and assuming $p > q$, we define the prolongation of $M$ as the separate prolongation of these image columns (as in Fig.~\ref{fig:prolongation}).
	The operation can therefore be expressed as
	\begin{equation}
	P \begin{bmatrix}
	M\\M\\M\\M
	\end{bmatrix} \in \R^{4 p \times q},
	\label{eq:matrix_prolongation}
	\end{equation}
	where $P \in \R^{4 p \times 4 p}$ is a suitable permutation.
	
	As the singular values of a matrix are invariant under row-permutations (see \eqref{eq:svd_permutation}), one can however restrict the analysis of the singular values for \eqref{eq:matrix_prolongation} to the case $P = I$.
	Let an \textit{economic SVD} of $M$ now be given by $M = U \Sigma V^{\top}$ with $U \in \R^{p \times q}$ and $\Sigma \in \R^{p \times p}$ \cite[Chapter 2.5]{Golub1996}.
	Then it holds
	\begin{equation}
	\underbrace{\begin{bmatrix}
		M\\M\\M\\M
		\end{bmatrix}}_{=: \hat{M}} =
	\underbrace{\begin{bmatrix}
		U\\U\\U\\U
		\end{bmatrix}}_{=: \hat{U}} \Sigma V^{\top}.
	\label{eq:svd_prolonged}
	\end{equation}
	\eqref{eq:svd_prolonged} however does not constitute a valid (economic) SVD of $\hat{M}$, since the columns of $\hat{U}$ are no longer normalized:
	One has $(\hat{U}^{\top} \hat{U})_{i, j} = 4$ for $i = j$ and $(\hat{U}^{\top} \hat{U})_{i, j} = 0$ for $i \neq j$.
	In order to regain a valid SVD, a factor of $2$ has to be redistributed from $\hat{U}$ to $\Sigma$, i.e.,
	\begin{equation}
	\hat{M} = (\hat{U} / 2) (2 \Sigma) V^{\top}.
	\end{equation}
	This implies $||\hat{M}||_* = 2 ||M||_*$, which in turn implies that the sought factor is given by $2$.
	Also note that this result can easily be generalized to the case of $d$-dimensional images, where that factor is given $2^{(d / 2)}$.
	
	The overall solution scheme is summarized by Alg.~\ref{alg:ml}, where an image domain of $\Omega = [0, m] \times [0, n]$ is assumed for the input images $T_1, \ldots, T_N \in \R^{m \times n}$.
	Further assumed are a predefined number $n_{lev}$ of resolution stages, predefined numbers $n_{iter}^j$ of linearization steps per stage (for $j = 1, \ldots, n_{lev}$), as well as a final relative threshold parameter of $\beta = \exp(\ln(\alpha) \sum_{j = 1}^{n_{lev}} n^j_{iter})$ (see subsection \ref{subsec:linearization}).
	
	\section{Data}
	\label{sec:data}
	
	\subsection{Synthetic Dataset: Textured Ellipse}
	\label{subsec:ellipse}
	The purpose of the first synthetic dataset is to illustrate the capacity of our model to correct the motion of objects with recurring changes in texture, exposing the inherent low-dimensional structure of the dataset.
	
	The image sequence is comprised of ten frames displaying a textured ellipse moving in a semicircular manner before a black background that further features a fixed white rectangle and a fixed white frame.
	The texture of the ellipse alternates between vertical stripes for all oddly indexed frames and horizontal stripes for all evenly indexed frames.
	
\begin{figure}[b]
		\centering
		\resizebox*{.65\columnwidth}{!}{
			\begin{tikzpicture}[every text node part/.style={align=center}]
			\node[inner sep=0pt, text width=2cm] at (0, 0) {$T_1$ \\[.25em] \includegraphics[width=2cm]{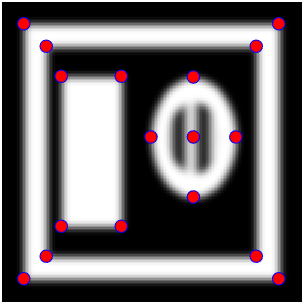}};
			\node[inner sep=0pt, text width=2cm] at (2.25, 0) {$T_4$ \\[.25em] \includegraphics[width=2cm]{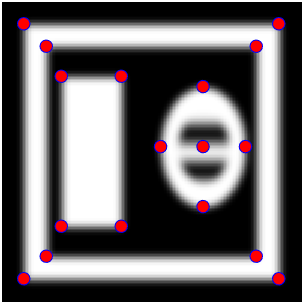}};
			\node[inner sep=0pt, text width=2cm] at (4.5, 0) {$T_7$ \\[.25em] \includegraphics[width=2cm]{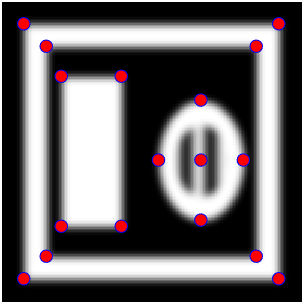}};
			\node[inner sep=0pt, text width=2cm] at (6.75, 0) {$T_{10}$ \\[.25em] \includegraphics[width=2cm]{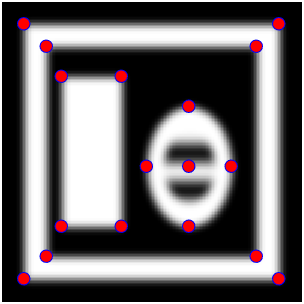}};
			\end{tikzpicture}
		}
		\caption{
			Exemplary frames from the textured ellipse-dataset with their respective landmarks.
			A perfect motion correction is expected to unify the ellipse positions, while keeping the white frames and rectangles stationary
		}
		\label{fig:textured_ellipse_input}
	\end{figure}

	To quantify the accuracy of registration on this dataset, we equipped each frame with 17 landmarks at the same corresponding (analytically determined) positions.
	All frames were generated at a resolution of $200 \times 200$ pixels.
	As an example, four out of the ten input frames are displayed along with their landmarks in Fig.~\ref{fig:textured_ellipse_input}.
	
	\subsection{Real-world Dataset I: Cardiac MRI}
	\label{subsec:heart}
	Besides the challenge of motion correction in the presence of recurring changes in object appearance that the first synthetic dataset posed, the first real-world dataset comes with the additional difficulty of irregular disturbances to object appearance.
	
	The sequence consists of cardiac MRI data in the so-called two-chamber view, where the left atrium and ventricle are on display.
	Seven repetitions of the heart cycle with blood flow in and out of the two chambers as well as breathing-induced motions of several structures like the thorax, the diaphragm, and the heart are shown.
	
	For this dataset, changes in object appearance relate to different phases of the heart cycle as we selected one frame from each systole, one frame from each diastolic relaxation and one frame from each diastolic filling (making for a total of 21 input frames).
	Due to the turbulent nature of the blood flow, the visual appearances of these phases are somewhat irregular and pose an interesting test case for the low-rank/sparse decomposition generated by our model.
	
\begin{figure}[t]
		\centering
		\resizebox*{.65\columnwidth}{!}{
			\begin{tikzpicture}[every text node part/.style={align=center}]
			
			\node[above, rotate=90] at (-1.125, 0) {\tiny $1^{\mbox{st}}$ Heart Cycle~~};
			\node[inner sep=0pt, text width=2.25cm] at (0, 0) {\tiny Systole \\[.25em] \includegraphics[width=2.25cm]{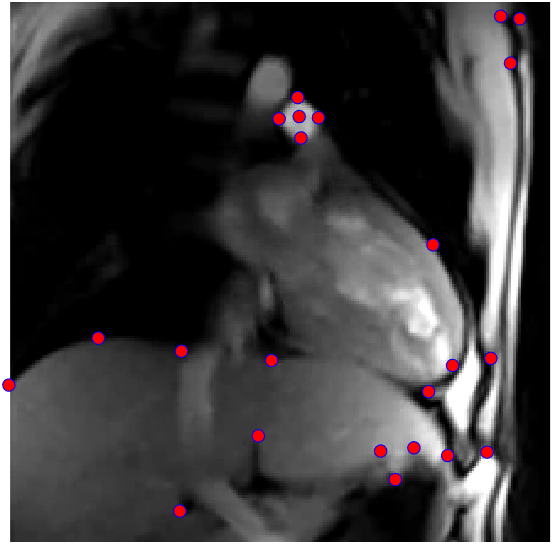}};
			\node[inner sep=0pt, text width=2.25cm] at (2.5, 0) {\tiny Diastolic Relaxation \\[.25em] \includegraphics[width=2.25cm]{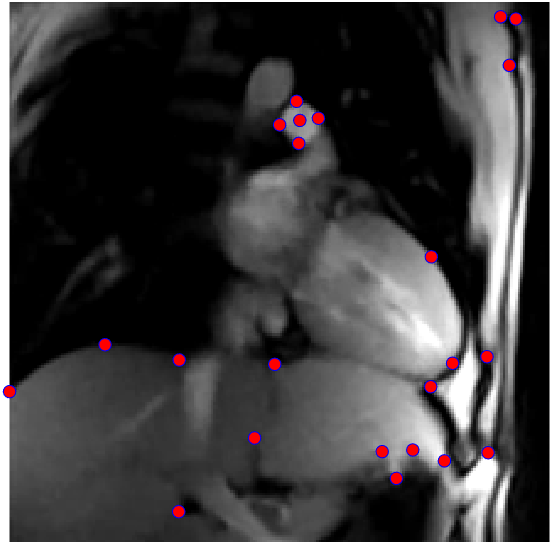}};
			\node[inner sep=0pt, text width=2.25cm] at (5, 0) {\tiny Diastolic Filling \\[.25em] \includegraphics[width=2.25cm]{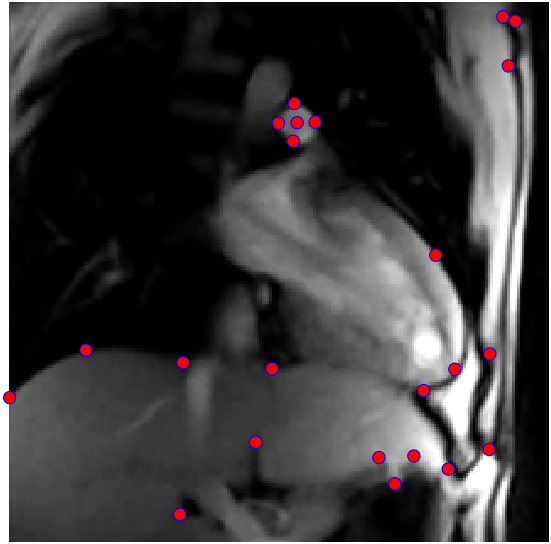}};
			
			\node at (0, -1.75) {\vdots};
			\node at (2.5, -1.75) {\vdots};
			\node at (5, -1.75) {\vdots};
			
			\node[inner sep=0pt, text width=2.25cm] at (0, -3.5) {\includegraphics[width=2.25cm]{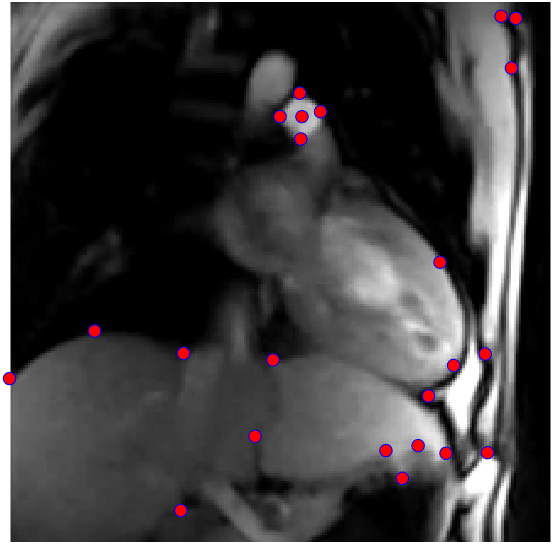}};
			\node[above, rotate=90] at (-1.125, -3.5) {\tiny $7^{\mbox{th}}$ Heart Cycle};
			\node[inner sep=0pt, text width=2.25cm] at (2.5, -3.5) {\includegraphics[width=2.25cm]{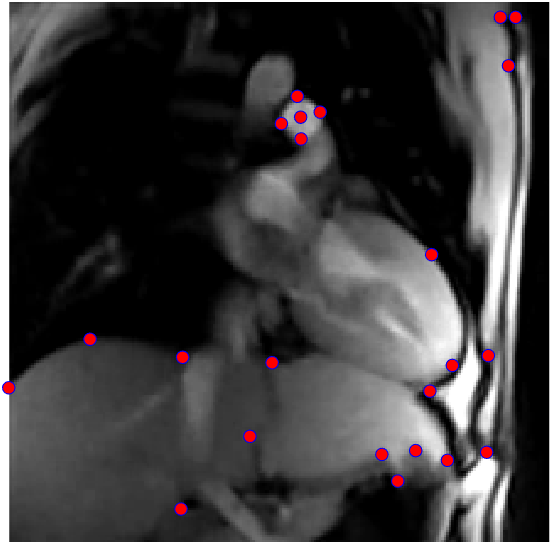}};
			\node[inner sep=0pt, text width=2.25cm] at (5, -3.5) {\includegraphics[width=2.25cm]{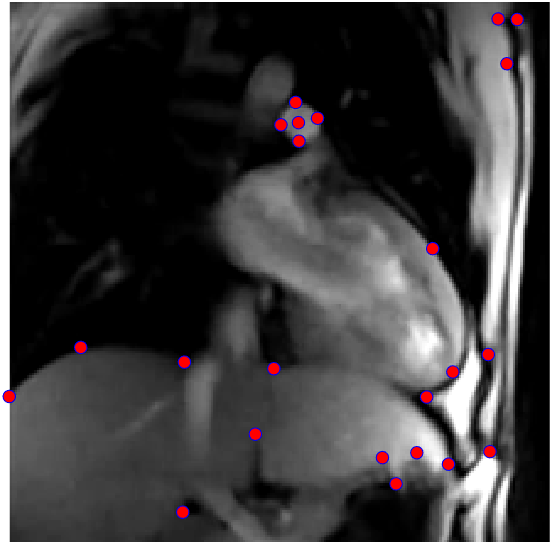}};
			
			\end{tikzpicture}
		}
		\caption{
			Exemplary frames from the cardiac MRI dataset with their respective landmarks.
			While clear visual congruences between different images of the same phase exist, they are obscured by the irregularity of the blood flow and pose a particular challenge to distance measures that model similarity as linear dependence
		}
		\label{fig:cardiac_input}
	\end{figure}
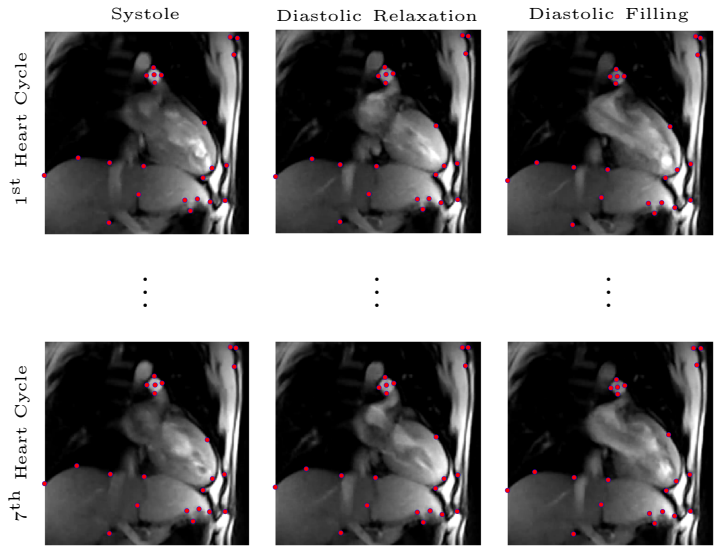

	As with the textured ellipse-dataset, we equipped this sequence with 23 handselected landmarks per frame.
	Each individual image was resolved with $220 \times 220$ pixels.
	The respective input frames for the first and last heart cycle are displayed together with their respective landmarks in Fig.~\ref{fig:cardiac_input}.
	
	\subsection{Real-world Dataset II: Challenging Data for Stereo and Optical Flow}
	\label{subsec:cdsof}
	
	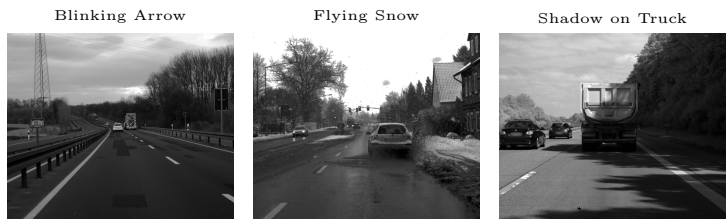
\begin{figure}[t]
		\centering
		\resizebox*{.65\columnwidth}{!}{
			\begin{tikzpicture}
			
			\node[inner sep=0pt] at (0, 0) {\includegraphics[width=2.75cm]{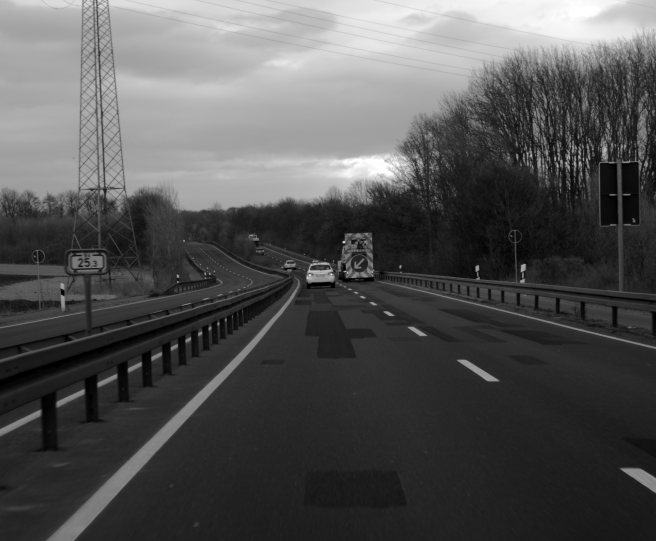}};
			\node[above] at (0, 1.15) {\tiny Blinking Arrow};
			\node[inner sep=0pt] at (3, 0) {\includegraphics[width=2.75cm]{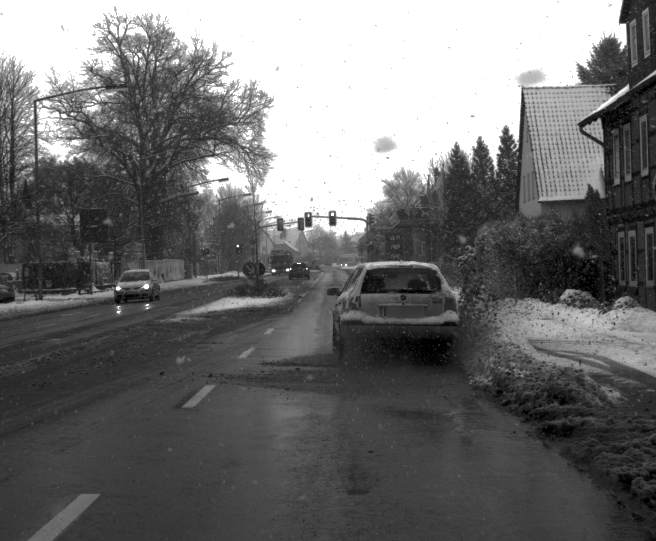}};
			\node[above] at (3, 1.15) {\tiny Flying Snow};
			\node[inner sep=0pt] at (6, 0) {\includegraphics[width=2.75cm]{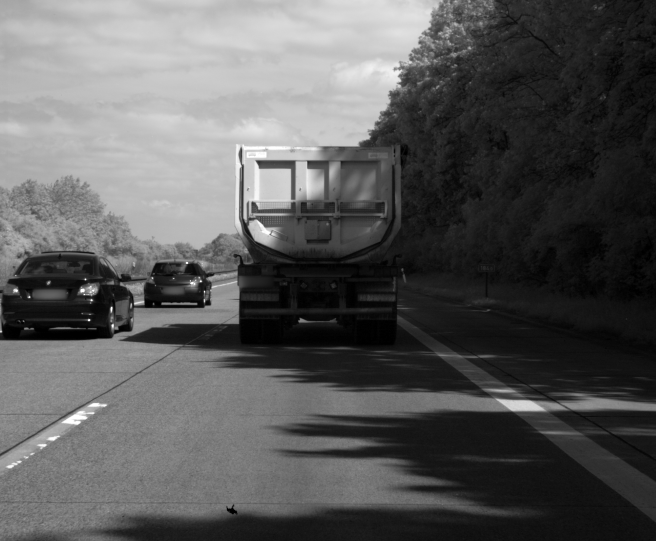}};
			\node[above] at (6, 1.15) {\tiny Shadow on Truck};
			
			\end{tikzpicture}
		}
		\caption{
			Reference frames for selected sequences from \cite{Meister2012}.
			The ``Blinking Arrow''-sequence \textbf{(left)} is deemed challenging because of intensity changes resulting from a blinking traffic sign, the challenge in the ``Flying Snow''-sequence \textbf{(center)} consists of heavy snowfall obstructing the view, and the ``Shadow on Truck''-sequence \textbf{(right)} features rapidly changing shadow patterns that do not describe physical motion
		}
		\label{fig:cdsof_input}
	\end{figure}
	
	We also evaluated our model on a variety of test sequences from the ``Challenging Data for Stereo and Optical Flow''-dataset (CDSOF) \cite{Meister2012}.
	This dataset features eleven sequences captured in real-world traffic situations that are deemed challenging for motion estimation algorithms due to diverse phenomena such as illumination changes from blinking signs, occlusions from snowflakes, and blurs from water spray.
	
	We selected subsequences of $10$ frames from the datasets entitled ``Blinking Arrow'', ``Flying Snow'' and ``Shadow on Truck'' as test cases as they all feature different distortions that pose interesting challenges to the robustness of our model.
	In order to restrict the required computational effort, we downsampled all used frames to a resolution of $271 \times 328$ pixels.
	
	Contrary to the other two datasets, all sequences from \cite{Meister2012} come with a predefined reference frame, which is why we drop the uniqueness constraint from section~\ref{subsec:uniqueness} for these inputs.
	Instead, we enforce alignments with the reference through a constraint of the form $\delta_{\{ 0 \}}( u^{(ref)} )$, where $u^{(ref)}$ is the displacement field for the respective reference.
	The reference images for all selected sequences are shown in Fig.~\ref{fig:cdsof_input}.

	\section{Results}
	\label{sec:results}
	For the former two datasets from section~\ref{sec:data}, we compare our registration approach to the following two methods:
	\begin{enumerate}
		\item An approach based on the simple variance dissimilarity measure given by
		\begin{equation}
		D_{\text{VAR}} (T_1, \ldots, T_N) := \frac{1}{2} \sum_{k = 1}^N || T_k - \bar{T} ||_2^2  \mbox{ with } \bar{T} = \sum_{k = 1}^N \frac{T_k}{N},
		\label{eq:var_measure}
		\end{equation}
		which has previously been used by \cite{Bhatia2007,Metz2011}.
		We combine \eqref{eq:var_measure} with the same TV-regularization and the same uniqueness constraint as in our model \eqref{eq:full_model}.
		\item A publicly available implementation of the $D_{\text{PCA2}}$-metric from \cite{Huizinga2016} in the {\ttfamily elastix} software package \cite{Klein2010}.
		This method uses a cubic B-spline transformation model and implicit regularization.
	\end{enumerate}
	For each individual landmark, accuracy is measured in terms of mean Euclidean distance to the mean landmark position
	\begin{equation}
	\frac{1}{N} \sum_{k = 1}^N || y_i^k - \bar{y}_i ||_2 \quad \mbox{with} \quad \bar{y}_i := \sum_{k = 1}^N \frac{y_i^k}{N}.
	\label{eq:lm_acc}
	\end{equation}
	$y_i^k \in \R^2$ therein denotes the position of the $i$-th landmark in the $k$-th image.
	
	For the CDSOF-datasets we compared our approach to the publicly available implementation of the ``nonlocal'' optical flow estimation method from \cite{Sun2010}, that was suggested as a referemce by the authors of \cite{Meister2012}.
	\cite{Sun2010} extends the classical Horn-Schunck model for optical flow estimation \cite{Horn1981} by a number of techniques, e.g., an additional nonlocal term derived from median filtering.
	To give a meaningful comparison, we altered our algorithm to include the median filtering of flow fields in between linearization steps as well.
	Note that the reference method only operates in a pairwise fashion.
	
	All experiments were performed on an Intel Core i7-8700 ($6 \times$ 3.20 GHz) system with 64~GB of memory, running {\scshape Matlab} R2019a under Ubuntu 18.04 (64-Bit).
	Our implementation is publicly available at
	\begin{center}
		\url{https://github.com/roland1993/d_RPCA}.
	\end{center}
	Computation times for the textured ellipse- and cardiac MRI-datasets are given in Tab.~\ref{tab:times} and for the CDSOF-dataset in Tab.~\ref{tab:times_cdsof}.
	
	\begin{table}[b]
		\centering
		\renewcommand{\arraystretch}{1.25}
		\resizebox{.5\linewidth}{!}{
			\begin{tabular}{|l | c c c|}
				\hline
				& $D_{\dRPCA} + \TV$ & $D_{\text{VAR}} + \TV$ & $D_{\text{PCA2}}$\\
				\hline
				Textured Ellipse & 12m\,0s & 6m\,27s & 40m\,4s\\
				Cardiac MRI & 39m\,8s & 29m\,20 & 1h\,54m\,11s\\
				\hline
			\end{tabular}
		}
		\caption{Computation times of competing methods}
		\label{tab:times}
	\end{table}

	\subsection{Textured Ellipse}
	\label{subsec:res_ellipse}
	
\begin{figure}[p]
		\centering
		\resizebox{.95\textwidth}{!}{
			\begin{tikzpicture}
			
			\node[rotate=90, align=center] at (-2.5, 0) {\bfseries Input + Deformation};
			\node[rotate=90, align=center] at (-2.5, -4.5) {\bfseries Output + Landmarks};
			\node[align=center] at (0, 2.25) {$T_1$};
			\node[align=center] at (5, 2.25) {$T_4$};
			\node[align=center] at (10, 2.25) {$T_7$};
			\node[align=center] at (15, 2.25) {$T_{10}$};
			
			\node[inner sep=0pt] at (0, 0) {\includegraphics[width=4cm]{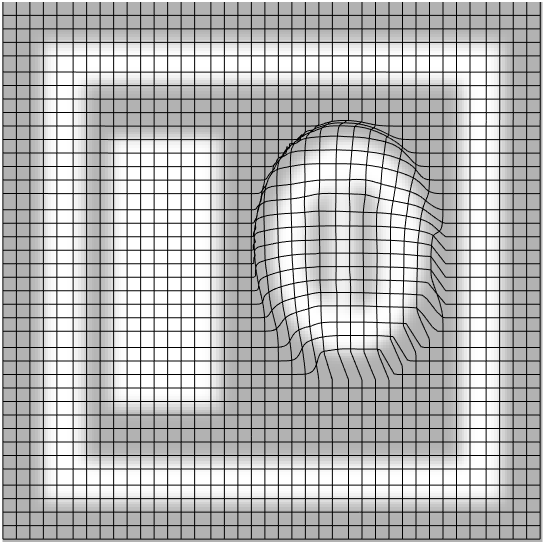}};
			\node[inner sep=0pt] at (5, 0) {\includegraphics[width=4cm]{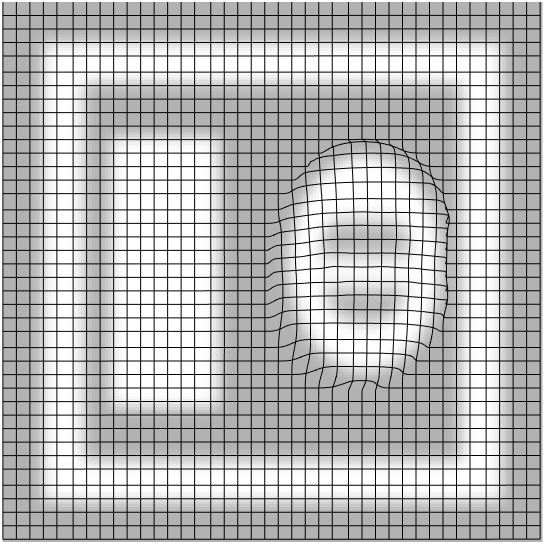}};
			\node[inner sep=0pt] at (10, 0) {\includegraphics[width=4cm]{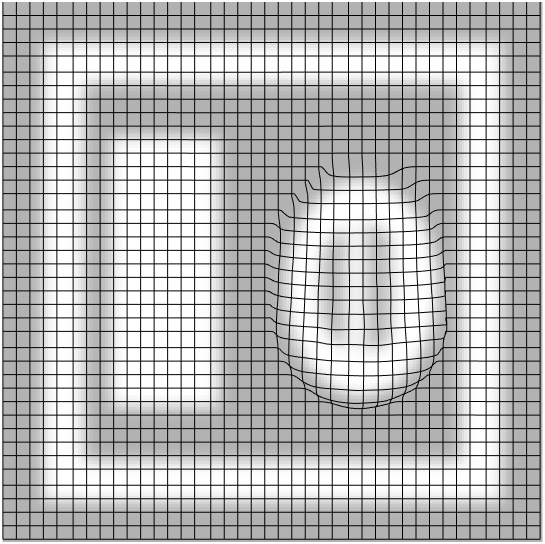}};
			\node[inner sep=0pt] at (15, 0) {\includegraphics[width=4cm]{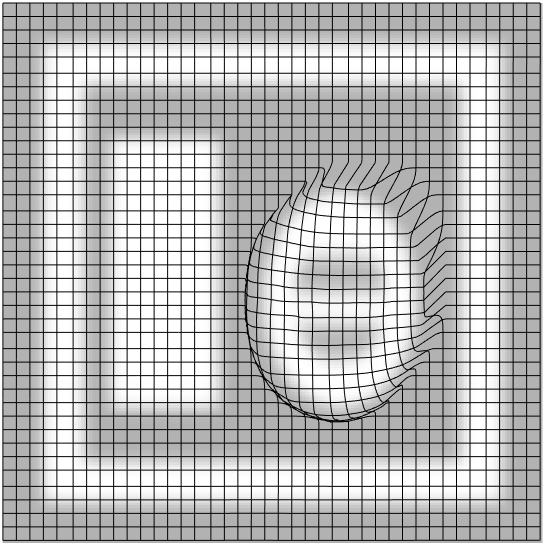}};
			
			\node[inner sep=0pt] at (0, -4.5) {\includegraphics[width=4cm]{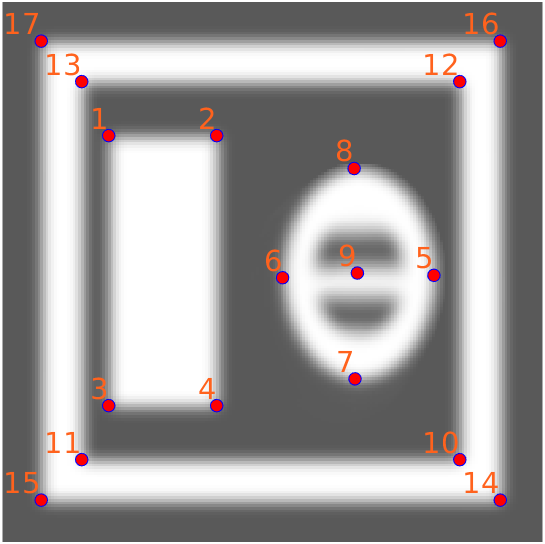}};
			\node[inner sep=0pt] at (5, -4.5) {\includegraphics[width=4cm]{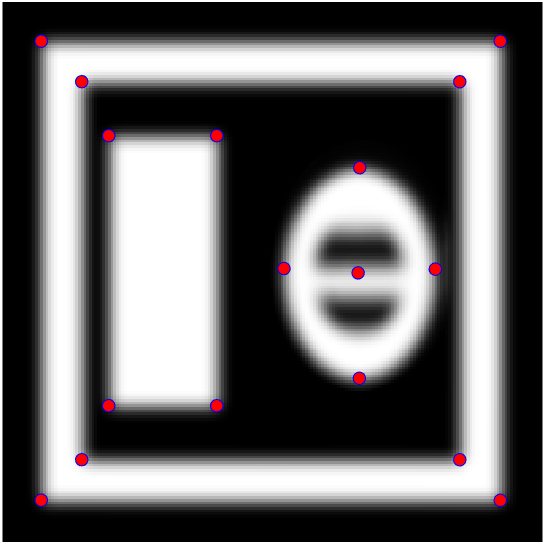}};
			\node[inner sep=0pt] at (10, -4.5) {\includegraphics[width=4cm]{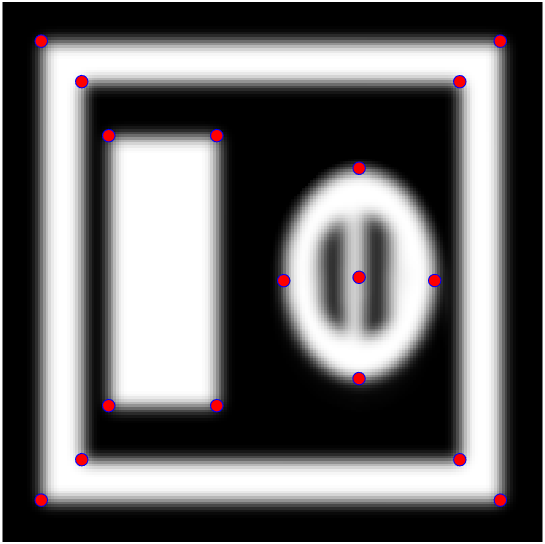}};
			\node[inner sep=0pt] at (15, -4.5) {\includegraphics[width=4cm]{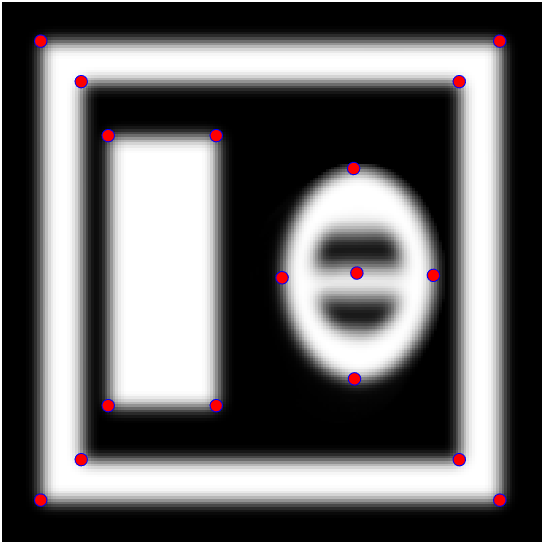}};

			\end{tikzpicture}
		}
		\caption{Results of proposed approach for the selected frames from Fig.~\ref{fig:textured_ellipse_input}.
			Our model allows to correct the motion of the ellipse through piecewise constant deformations \textbf{(top row)}, while automatically detecting and discarding repetitive structural noise (the horizontal and vertical bars) in the registration process.
			The motion-corrected images \textbf{(bottom row)} exhibit a good visual correspondence between matching landmarks.
			Quantitative results are given in Fig.~\ref{fig:ellipse_lm_accuracy}, in which landmarks are indexed in the same order as in $T_1$
		}
		\label{fig:ellipse_results}
	\end{figure}
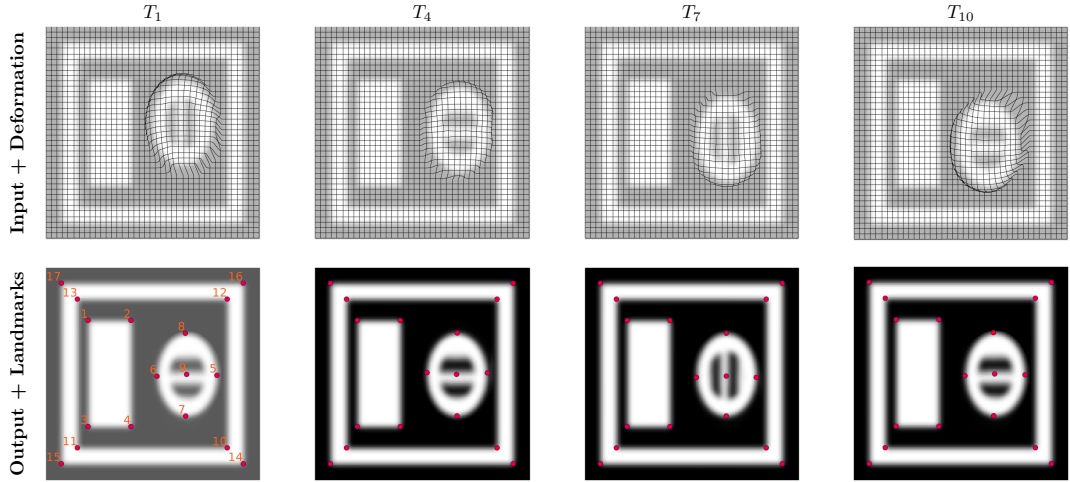

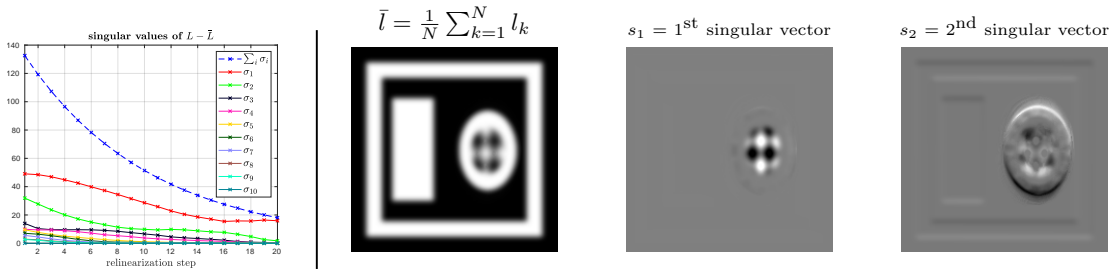
\begin{figure}[p]
		\centering
		\resizebox{\textwidth}{!}{
			\begin{tikzpicture}
			
			\node at (0, 0) {\includegraphics[width=4cm]{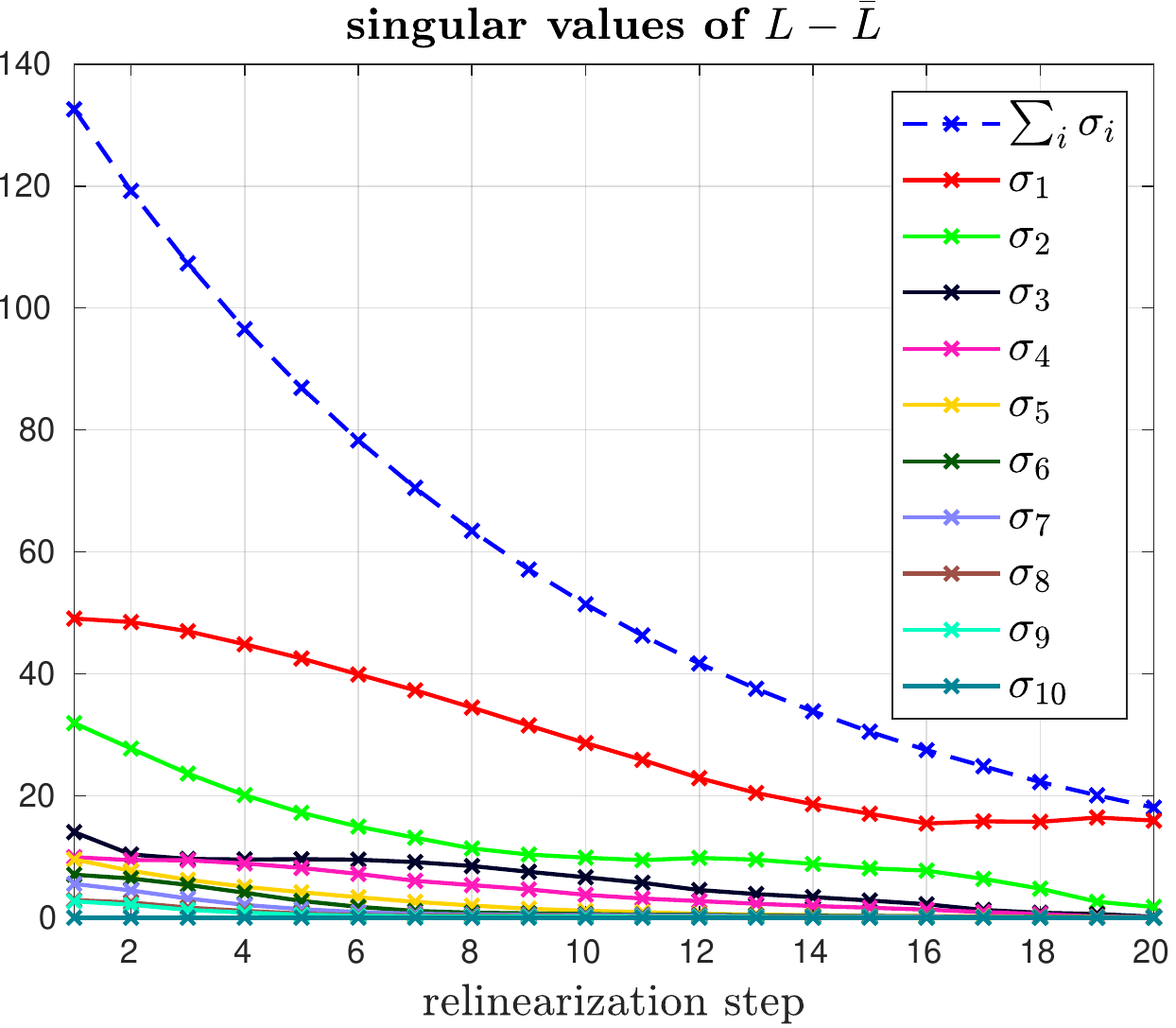}};
			\draw[thick] (2.5, -1.75) -- (2.5, 1.75);
			
			\node at (4.5, 0) {\includegraphics[width=3cm]{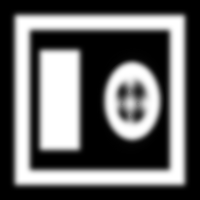}};
			\node[above] at (4.5, 1.5) {$\bar{l} =  \frac{1}{N} \sum_{k = 1}^N l_k$};
			\node at (8.5, 0) {\includegraphics[width=3cm]{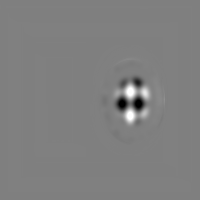}};
			\node[above] at (8.5, 1.5) {\scriptsize $s_1$ = $1^{\mbox{st}}$ singular vector};
			\node at (12.5, 0) {\includegraphics[width=3cm]{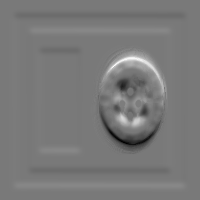}};
			\node[above] at (12.5, 1.5) {\scriptsize $s_2$ = $2^{\mbox{nd}}$ singular vector};
			
			\end{tikzpicture}
		}
		\caption{Singular values and vectors of centered low-rank components $L - \bar{L}$.
			The singular value progression \textbf{(left)} over the inner iterations of Alg.~\ref{alg:ml} (scaling adjusted, see subsection~\ref{subsec:multilevel}) shows that the norm $|| L - \bar{L} ||_*$ -- the quantity constrained by our model -- is dominated by the singular value $\sigma_1$ towards the end of the iteration.
			This indicates that the columns of $L$ can be reconstructed from their mean $\bar{l}$ \textbf{(second from left)} and the dominating singular vector $s_1$ of the \textit{centered} low-rank components $L - \bar{L}$ \textbf{(second from right)} with only minor error.
			Intuitively, $s_1$ is added to $\bar{l}$ when reconstructing horizontal bars and subtracted from $\bar{l}$ when reconstructing vertical bars.
			In comparison, the second singular vector $s_2$ \textbf{(right)} only describes negligible remainders of motion with little influence on $L - \bar{L}$ (as indicated by the final magnitude of $\sigma_2$ in the left curve).
			To summarize, our method was able to automatically detect the low-dimensional texture variations present in the dataset
		}
		\label{fig:ellipse_sv}
	\end{figure}

	Using the parameters $\alpha = 0.9$, $\mu = 0.2$ as well as $n_{lev} = 3$ with $n_{iter}^1 = 16$ and $n_{iter}^j = 2$ for $j \geq 2$
	in Alg.~\ref{alg:ml}, we achieved the results displayed in \Cref{fig:ellipse_results,fig:ellipse_sv} for the textured ellipse-dataset.
	We shortly note that we observed values of $n_{iter}^j \geq 2$ for $j \geq 2$ to be mandatory in order for Alg.~\ref{alg:ml} to generate useful linearization points on higher resolution levels.
	Furthermore, a small enough $\alpha$ was crucial in finding low-rank components $L$ that accurately describe the structural changes in texture -- higher values of $\alpha$ on the other hand allowed for unwanted motion artifacts in $L$.
	
	For the variance method based on \eqref{eq:var_measure}, we employed an adapted version of Alg.~\ref{alg:ml} with the regularization strength set to $\mu = 0.1$.
	The number of resolution levels as well as the iterations per level were kept the same as for our method.
	
	In the \texttt{elastix}-based implementation of $D_{\text{PCA2}}$, we increased the settings recommended by the authors of \cite{Huizinga2016} to three resolution stages (instead of the recommended two), 2.000 iterations per stage (recommended: 1.000) and 25.000 random coordinates per stage (recommended: 2.048) in order to ensure sufficient computational capacity for the method to produce accurate solutions.
	
	Fig.~\ref{fig:ellipse_results} visualizes both the deformations calculated by our model and the warped images for all four exemplary frames from Fig.~\ref{fig:textured_ellipse_input}.
	Fig.~\ref{fig:ellipse_sv} further analyzes the generated low-rank components in terms of singular values and (left) singular vectors of $L - \bar{L}$, i.e., the matrix whose nuclear norm is constrained by our model \eqref{eq:full_model}.
	As the dataset is of synthetic nature and therefore without intensity distortions, the sparse outlier components $E = M - L$ are negligible and are not displayed.
	
\begin{figure}[p]
		\centering
		\includegraphics[width=.75\textwidth]{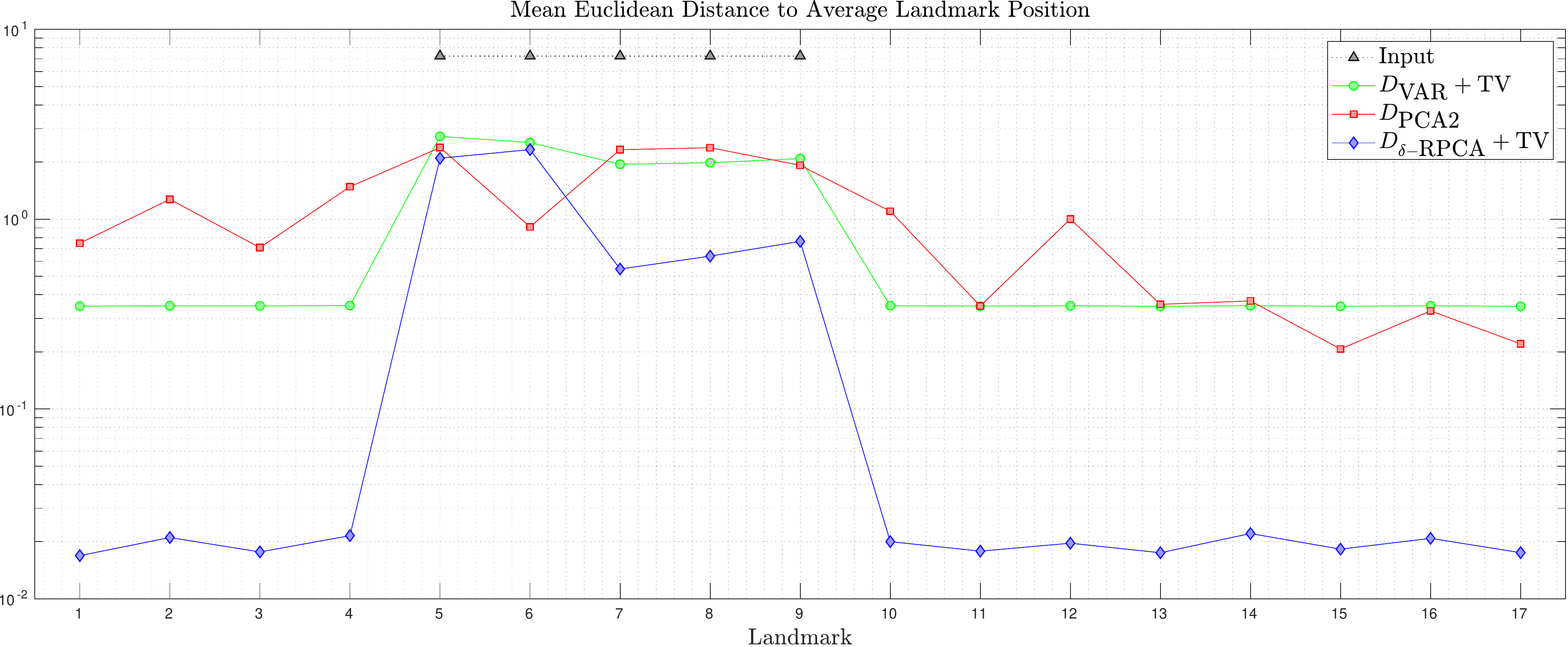}
		\caption{
			Comparison of landmark accuracy for the textured ellipse-dataset as measured by \eqref{eq:lm_acc} (lower is better).
			Landmarks are ordered as in Fig.~\ref{fig:ellipse_results} with landmarks  $5 - 9$ attributed to the moving ellipse and the remaining landmarks positioned around the stationary white rectangle and frame.
			Note that the latter do not appear in the input curve due to zero error and logarithmic axis scaling.
			Our method clearly outperforms the two competing approaches as it is able to correct the positions of the ``ellipse-landmarks'' most accurately without introducing artificial motion to the remaining landmarks
		}
		\label{fig:ellipse_lm_accuracy}
	\end{figure}

	A quantitative comparison in terms of landmark accuracy between our approach and the two competing methods is presented in Fig.~\ref{fig:ellipse_lm_accuracy}.
	The comparison shows that our method significantly outperforms the other approaches on this dataset:
	While it corrects the ellipse positions most accurately out of the three methods, it is also the only one that does not introduce notable motion to the white rectangle and frame (which were already stationary in the input sequence).
	
	In terms qualitative results, Fig.~\ref{fig:ellipse_sv} shows that our method was moreover able to find a near-perfect embedding of the motion-compensated images in a low-dimensional subspace:
	The negligible magnitudes of the singular values $\sigma_2, \ldots, \sigma_{10}$ of the centered low-rank components $L - \bar{L}$ indicate that a two-dimensional basis consisting of the mean low-rank component $\bar{l}$ and the singular vector $s_1$ is largely sufficient to approximate the output images.
	
	\subsection{Cardiac MRI}
	\label{subsec:res_heart}
	
	For the cardiac MRI dataset, we only adjusted the regularization strength and threshold-scaling parameters of our method to $\mu = 0.125$ and $\alpha = 0.95$.
	In the variance registration method, we kept all parameters fixed except for $\mu = 0.065$ and since the $D_{\text{PCA2}}$ model does not feature explicit regularization parameters, we left all the above settings unchanged for this method.
	
\begin{figure}[p]
		\centering
		\resizebox{1\textwidth}{!}{
			\begin{tikzpicture}
			
			\node[inner sep=0pt] at (0, 0) {\includegraphics[height=4cm]{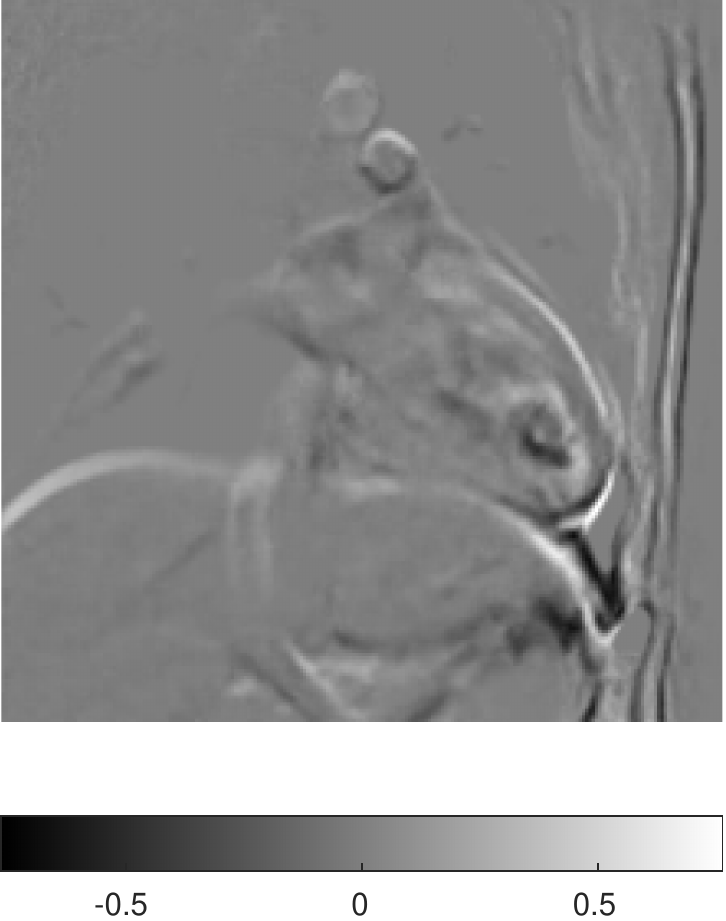}};
			\node[inner sep=0pt] at (4.5, 0) {\includegraphics[height=4cm]{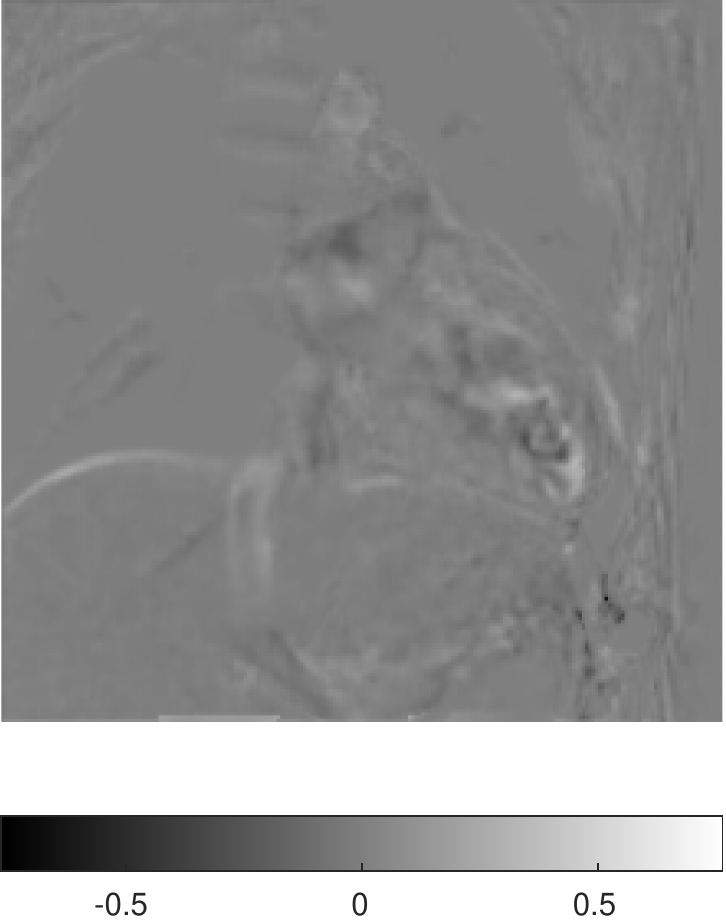}};
			
			\node[inner sep= 0pt] at (9.5, 0) {\includegraphics[width=4cm]{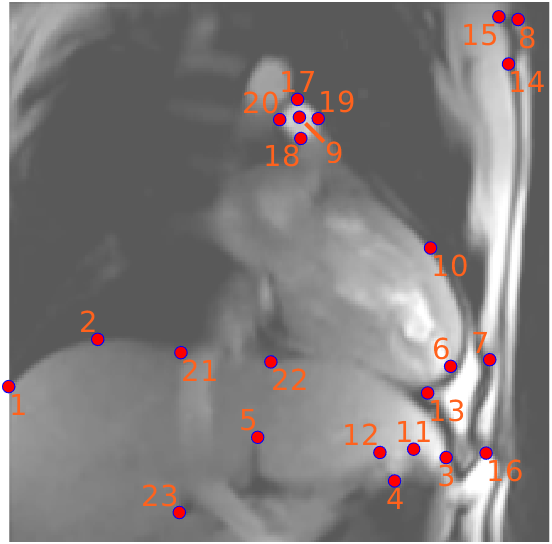}};
			\node[inner sep= 0pt] at (14, 0) {\includegraphics[width=4cm]{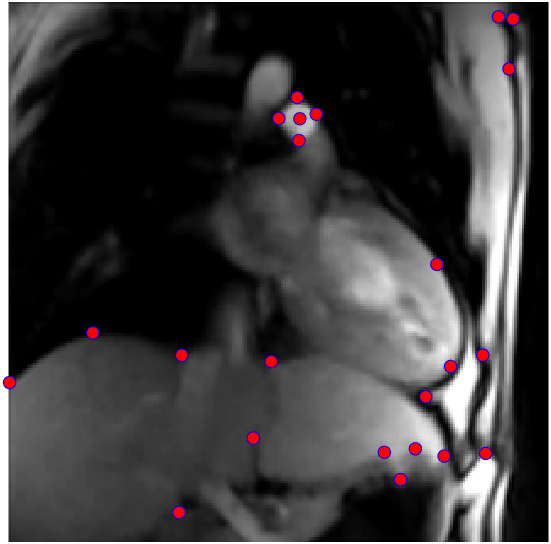}};
			
			\node[inner sep=0pt] at (0, -4.5) {\includegraphics[height=4cm]{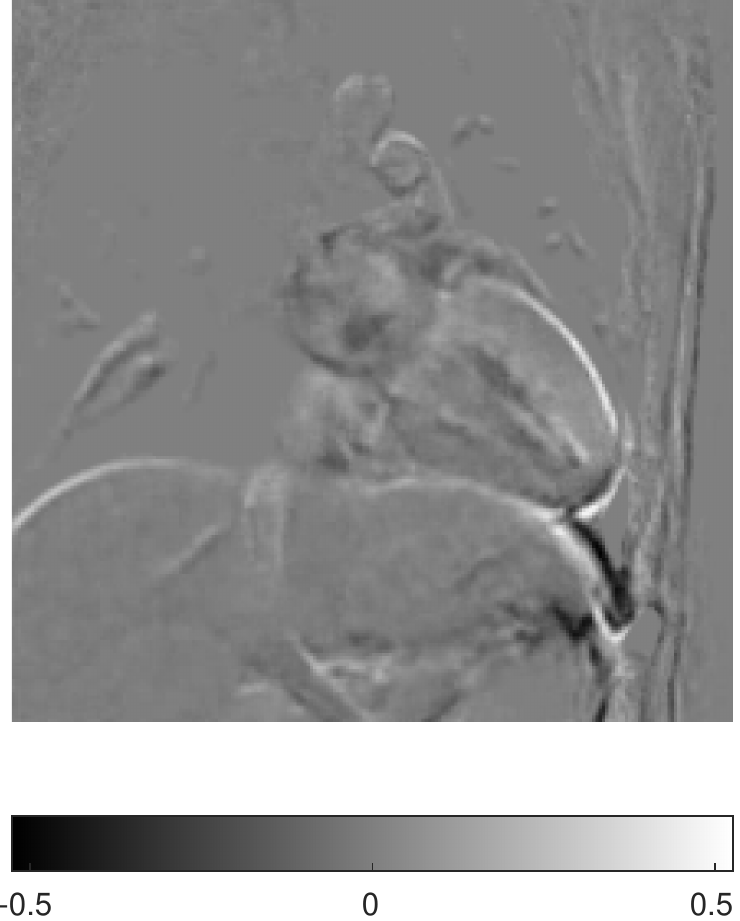}};
			\node[inner sep=0pt] at (4.5, -4.5) {\includegraphics[height=4cm]{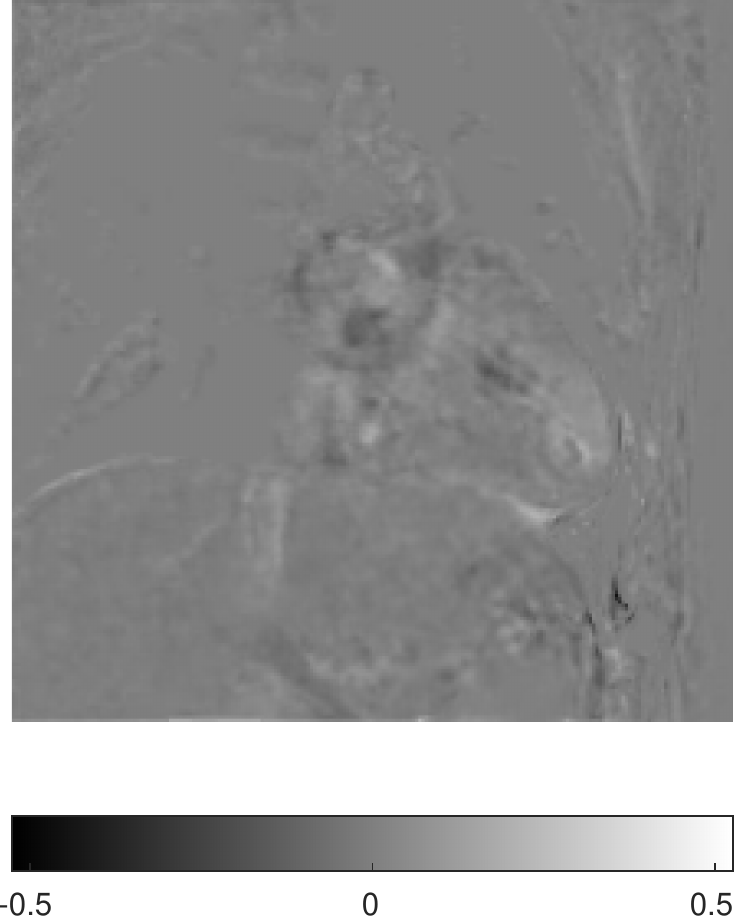}};
			
			\node[inner sep= 0pt] at (9.5, -4.5) {\includegraphics[width=4cm]{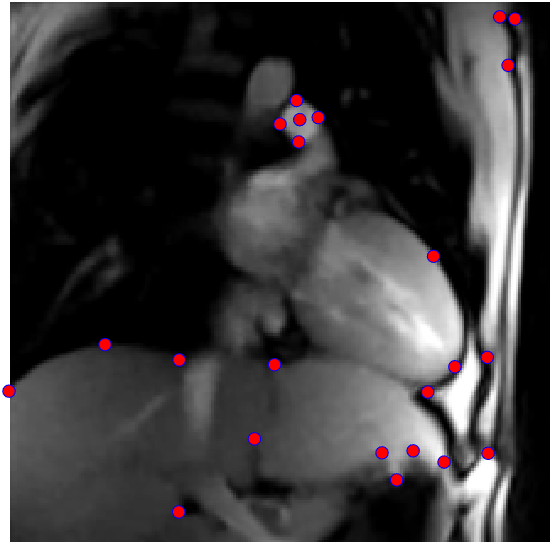}};
			\node[inner sep= 0pt] at (14, -4.5) {\includegraphics[width=4cm]{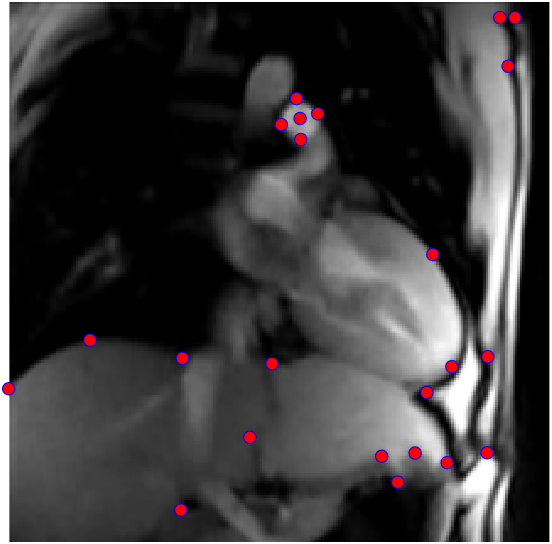}};
			
			\node[inner sep=0pt] at (0, -9) {\includegraphics[height=4cm]{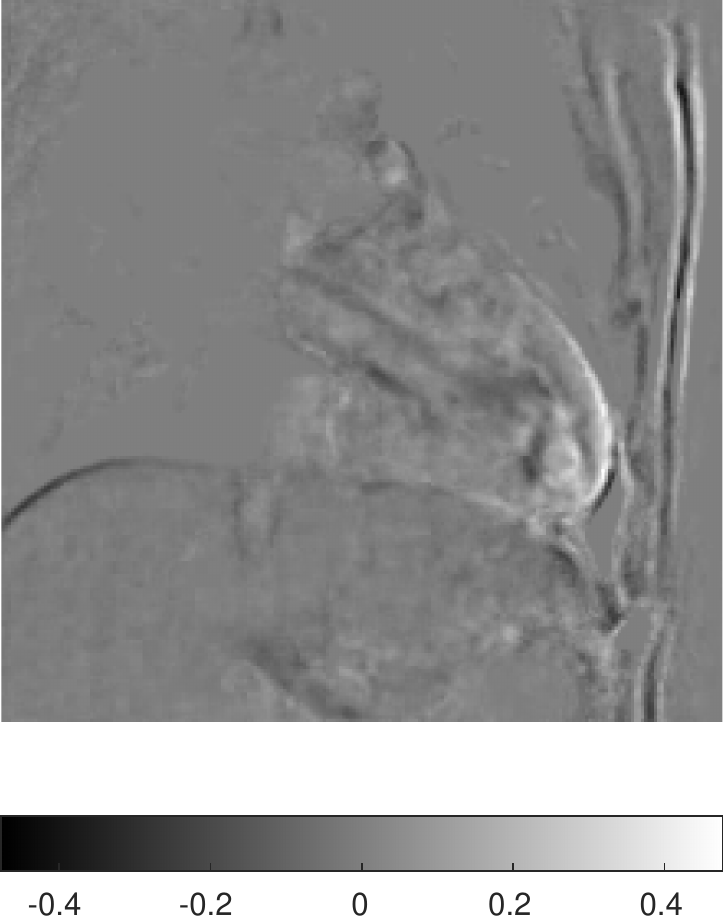}};
			\node[inner sep=0pt] at (4.5, -9) {\includegraphics[height=4cm]{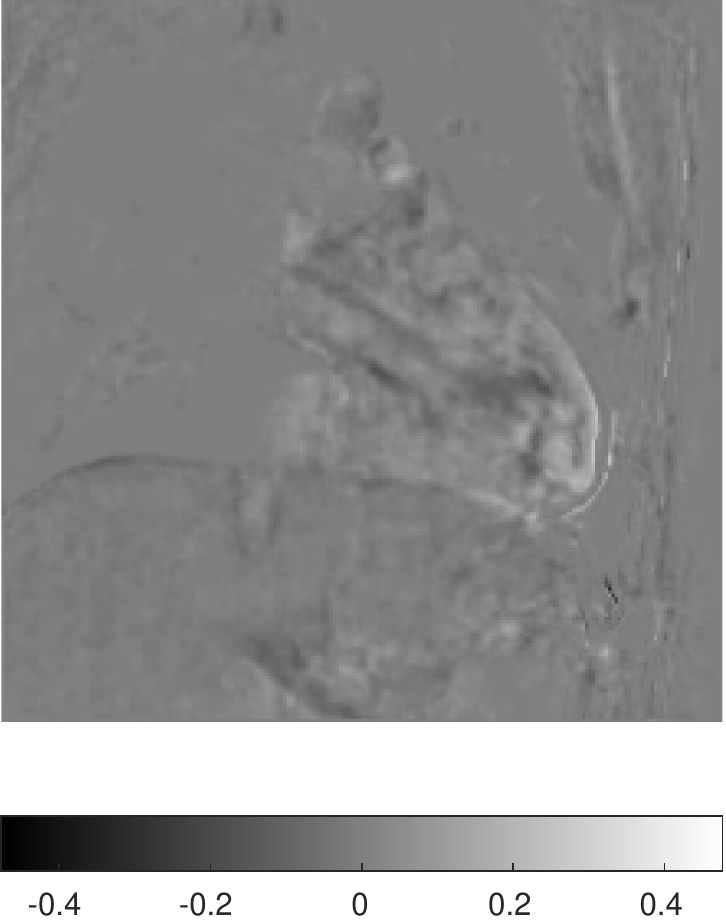}};
			
			\node[inner sep= 0pt] at (9.5, -9) {\includegraphics[width=4cm]{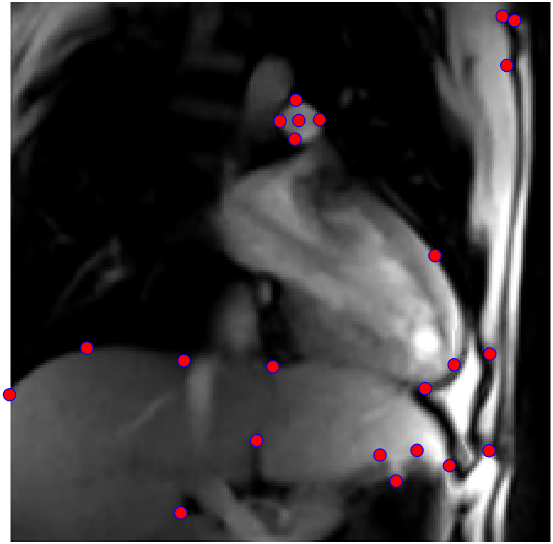}};
			\node[inner sep= 0pt] at (14, -9) {\includegraphics[width=4cm]{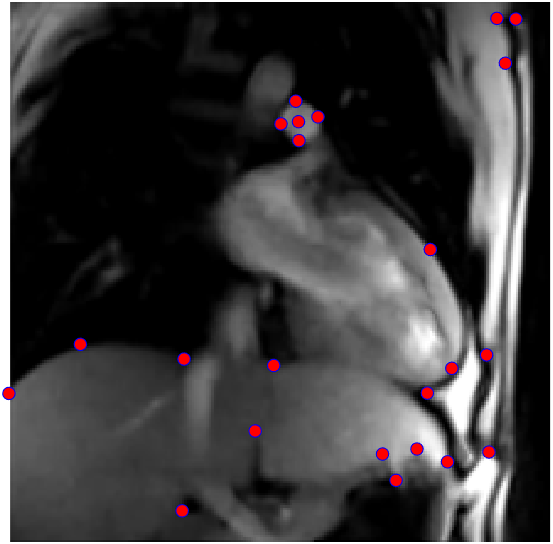}};

			\node[rotate=90, align=center] at (-2.5, 0) {\bfseries Systole};
			\node[rotate=90, align=center] at (-2.5, -4.5) {\bfseries Diastolic Relaxation};
			\node[rotate=90, align=center] at (-2.5, -9) {\bfseries Diastolic Filling};
			
			\node[align=center] at (0, 2.25) {Before Registration};
			\node[align=center] at (4.5, 2.25) {After Registration};
			\node[align=center] at (2.25, 2.75) {\textbf{Difference Images: $1^{\mbox{st}}$ vs. $7^{\mbox{th}}$ Heart Cycle}};
			
			\node[align=center] at (9.5, 2.25) {$1^{\mbox{st}}$ Heart Cycle};
			\node[align=center] at (14, 2.25) {$7^{\mbox{th}}$ Heart Cycle};
			\node[align=center] at (11.75, 2.75) {\textbf{Output + Landmarks}};
			
			\end{tikzpicture}
		}
		\caption{
			Results of proposed approach on the cardiac MRI-dataset.
			The difference images per phase between the first and seventh heart cycle before and after registration \textbf{(left two columns)} show a successful motion compensation:
			Motion in the thorax area as well as the front-facing area of the diaphragm and the heart apex is greatly reduced across all phases.
			The main intensity differences after registration are located inside the heart itself and are due to irregular turbulences of the blood flow.
			The two right-hand columns show the actual motion-compensated images including their deformed landmarks -- in the upper-left image, an additonal ordering of the landmarks is given that is referred to again in Fig.~\ref{fig:cardiac_lm_accuracy}
		}
		\label{fig:cardiac_results}
	\end{figure}

\begin{figure}[p]
		\centering
		\includegraphics[width=.75\textwidth]{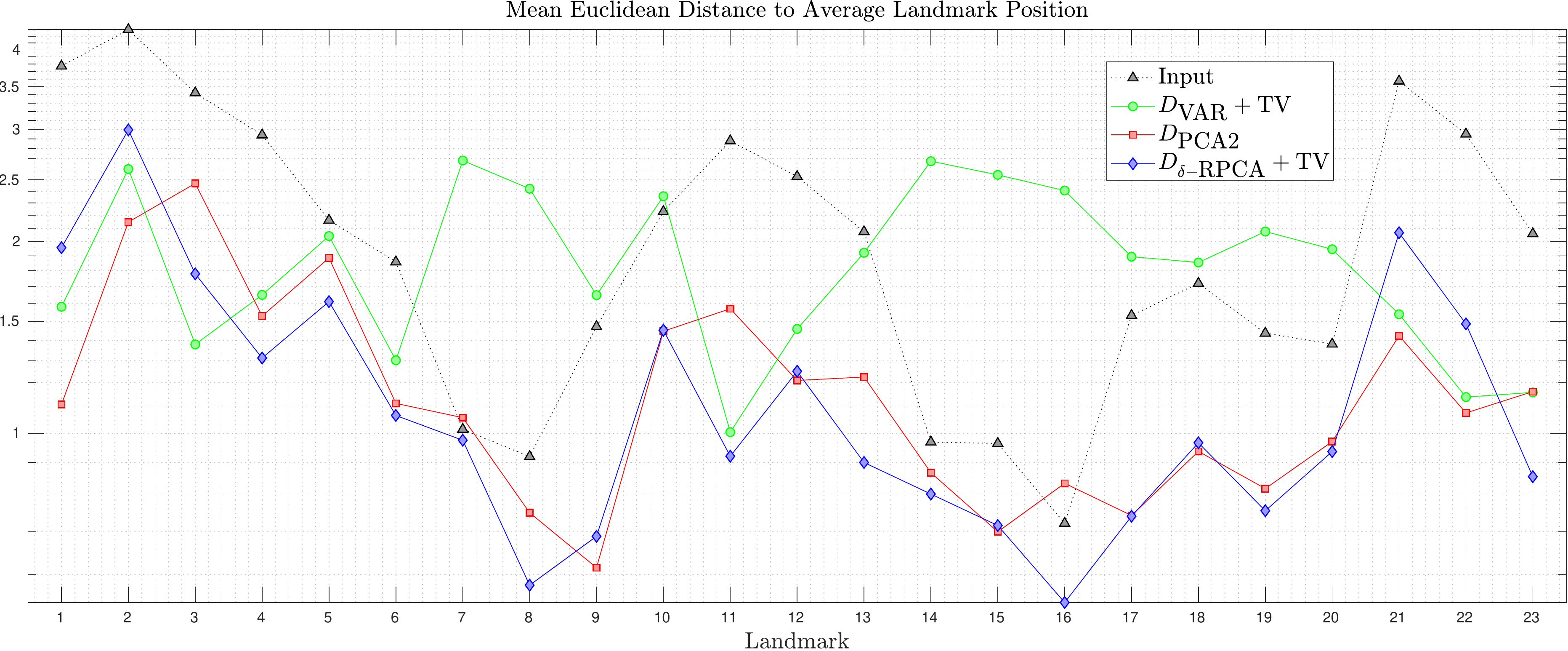}
		\caption{
			Comparison of landmark accuracy for the cardiac MRI-dataset as measured by \eqref{eq:lm_acc} (lower is better).
			Landmarks are ordered as in Fig.~\ref{fig:cardiac_results}.
			While our method still produces the most accurate deformations (performing best for 13 out of 23 landmarks), results are much more balanced than in the textured ellipse-experiment (see Fig.~\ref{fig:ellipse_lm_accuracy}) with $D_{\text{PCA2}}$ performing remarkably well despite being based on pure PCA.
			However, one notable drawback which both competing methods exhibit is that landmarks occasionally feature more motion in the registered images than was actually present in the input sequence -- this is indicated by crossings of their respective curves with the gray input curve (see for example landmarks 7 and 16 from the thorax area)
		}
		\label{fig:cardiac_lm_accuracy}
	\end{figure}

\begin{figure}[t]
		\resizebox{\textwidth}{!}{
			\begin{tikzpicture}
			
			\node at (0, 0) {\includegraphics[width=4cm]{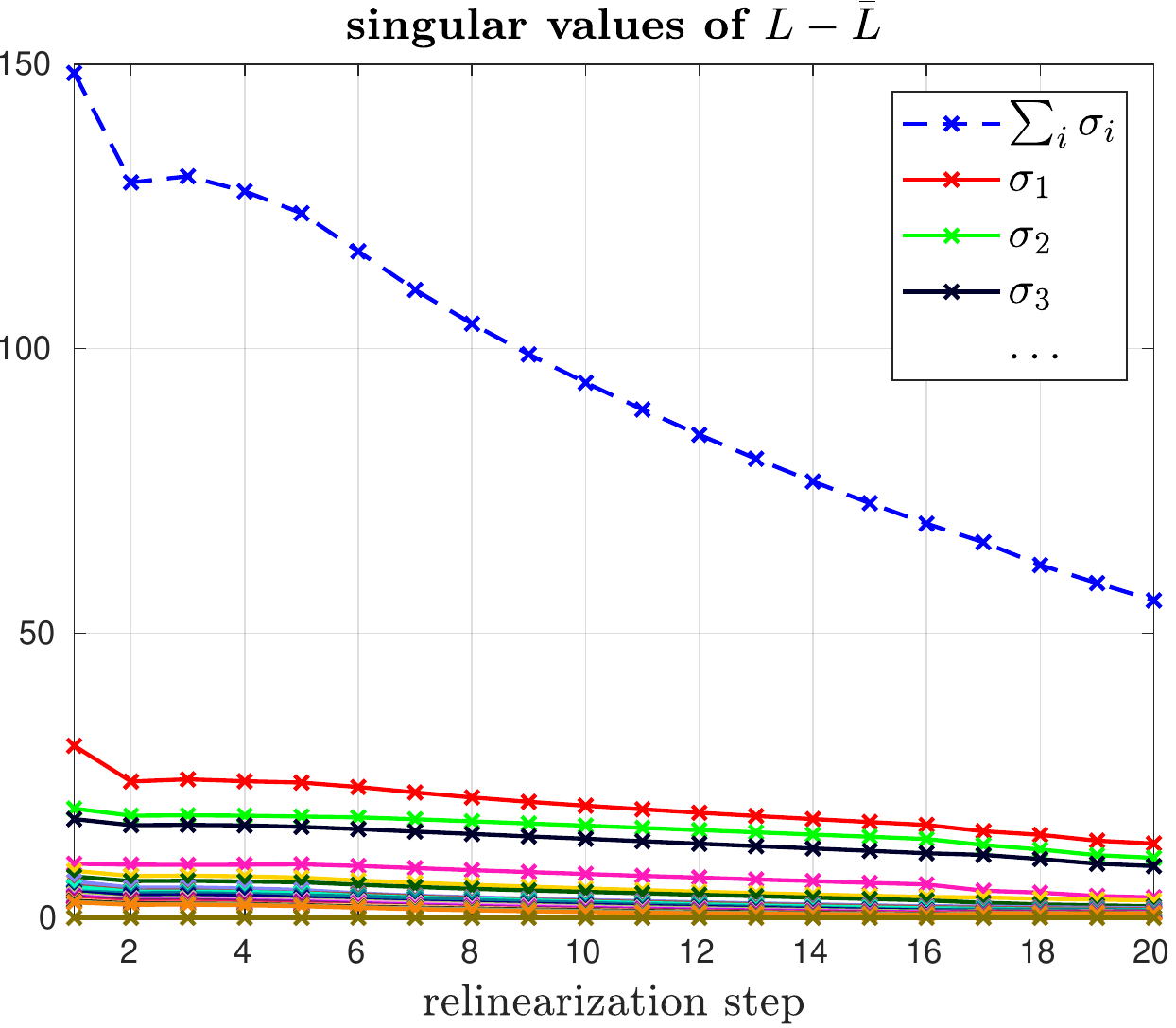}};
			\draw[thick] (2.5, -1.75) -- (2.5, 1.75);
			
			\node at (4.5, 0) {\includegraphics[width=3cm]{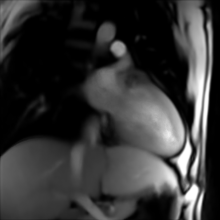}};
			\node[above] at (4.5, 1.5) {$\bar{l}$};
			\node at (8, 0) {\includegraphics[width=3cm]{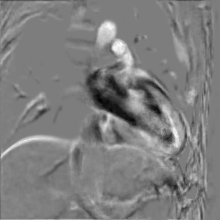}};
			\node[above] at (8, 1.5) {\scriptsize $s_1$};
			\node at (11.5, 0) {\includegraphics[width=3cm]{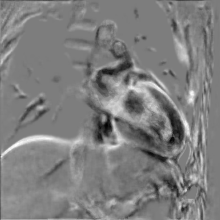}};
			\node[above] at (11.5, 1.5) {\scriptsize $s_2$};
			\node at (15, 0) {\includegraphics[width=3cm]{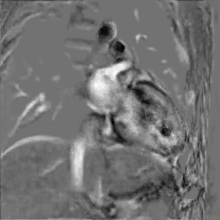}};
			\node[above] at (15, 1.5) {\scriptsize $s_3$};
			
			\end{tikzpicture}
		}
		\caption{
			Singular values and vectors of the centered low-rank components $L - \bar{L}$ for the cardiac MRI-dataset.
			The development of the singular values \textbf{(left)} shows, that the nuclear norm $||L - \bar{L}||_*$ (which is constrained by our model) is dominated by the three largest singular values $\sigma_1 - \sigma_3$.
			Consequently, the low-rank components of the warped images primarily consist of a linear combination of $\bar{l}$ \textbf{(second from left)} and the three corresponding singular vectors $s_1, s_2, s_3$ \textbf{(third, second, first from right)}.
			Especially note the visual congruences between these three singular vectors and the characteristics of the three considered heart phases:
			While $s_1$ marks a blood flow into the left ventricle (see the diastolic filling in Fig.~\ref{fig:cardiac_results}), $s_2$ highlights the mitral valve, which is clearly visibly closed during the diastolic relaxation phase (see again Fig.~\ref{fig:cardiac_results}).
			Moreover, $s_3$ exhibits a high contrast between atrium and ventricel as present during the systole in Fig.~\ref{fig:cardiac_results}
		}
		\label{fig:cardiac_sv}
	\end{figure}
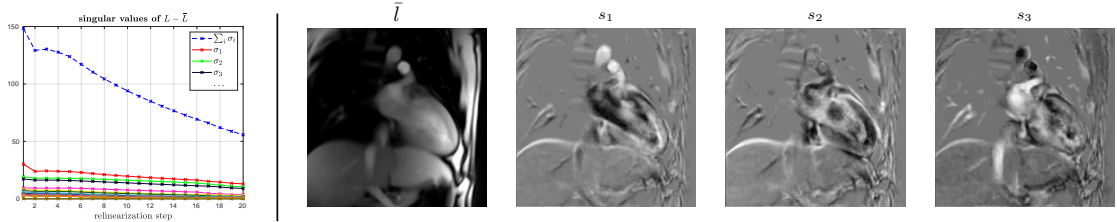

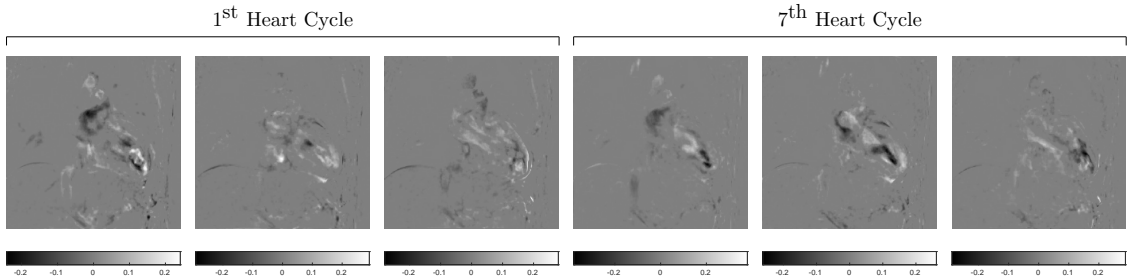
\begin{figure}[t]
		\centering
		\resizebox{\textwidth}{!}{
			\begin{tikzpicture}
			
			\draw (-1.5, 2.25) -- (8, 2.25) node[midway, above] {$1^{\mbox{st}}$ Heart Cycle};
			\draw (-1.5, 2.25) -- (-1.5, 2.1);
			\draw (8, 2.25) -- (8, 2.1);
			\draw (8.25, 2.25) -- (17.75, 2.25) node[midway, above] {$7^{\mbox{th}}$ Heart Cycle};
			\draw (8.25, 2.25) -- (8.25, 2.1);
			\draw (17.75, 2.25) -- (17.75, 2.1);
			
			\node[inner sep=0pt] at (0, 0) {\includegraphics[width=3cm]{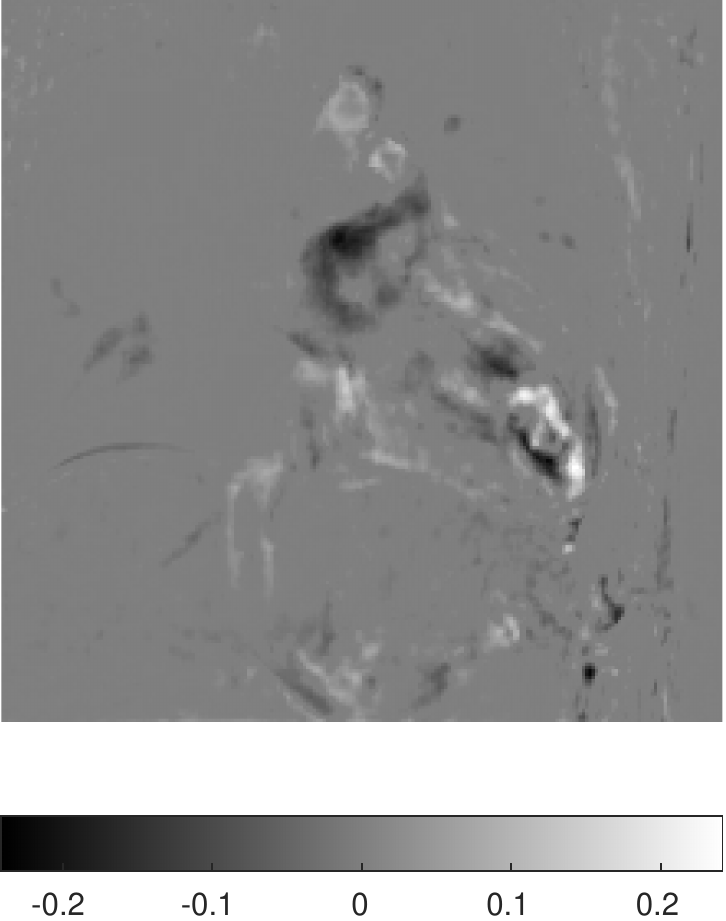}};
			\node[inner sep=0pt] at (3.25, 0) {\includegraphics[width=3cm]{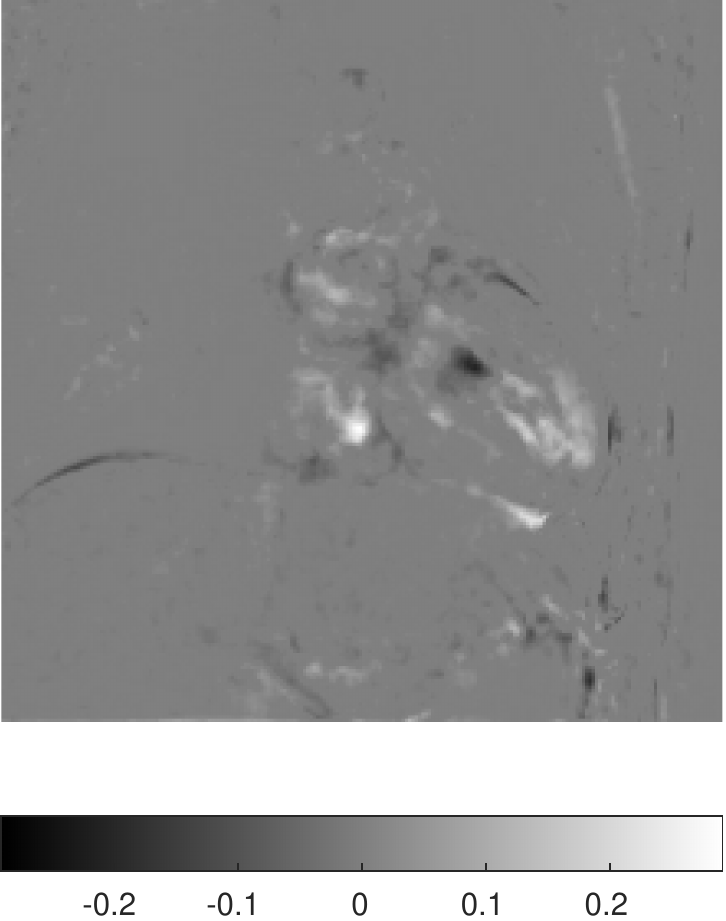}};
			\node[inner sep=0pt] at (6.5, 0) {\includegraphics[width=3cm]{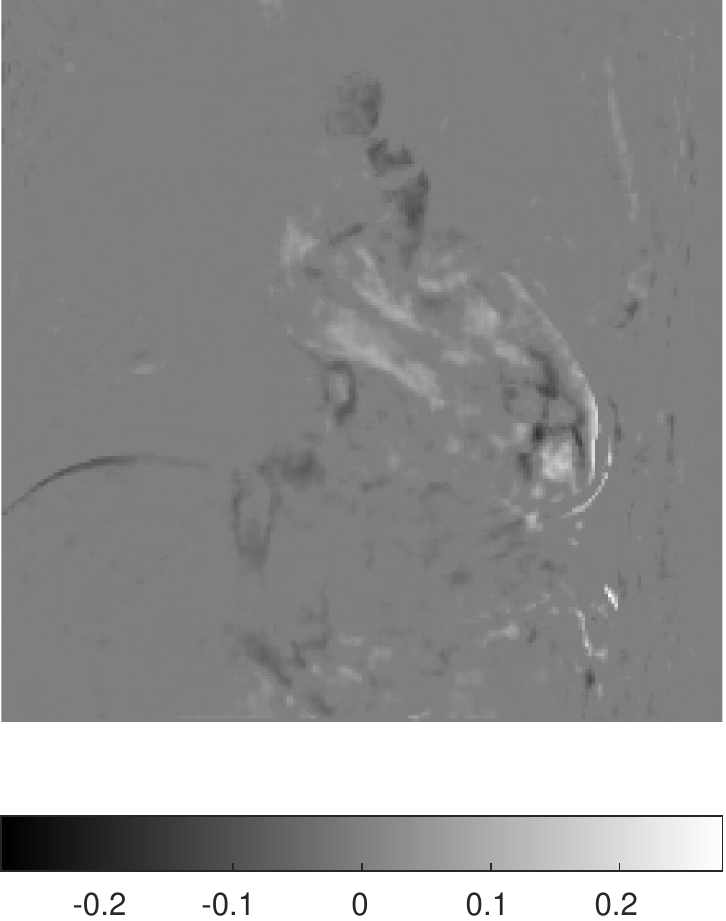}};

			\node[inner sep=0pt] at (9.75, 0) {\includegraphics[width=3cm]{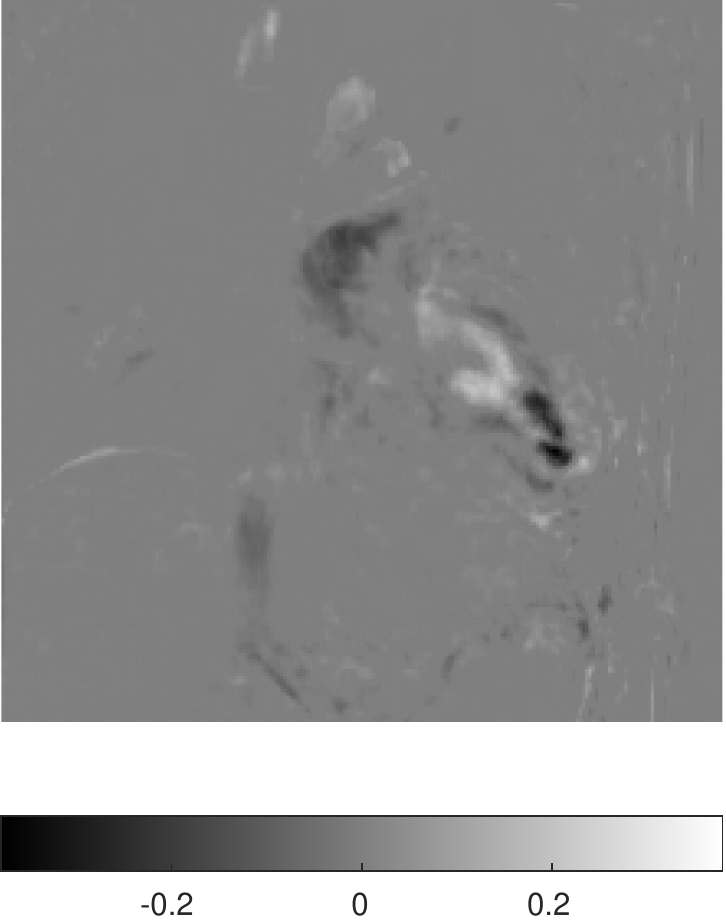}};
			\node[inner sep=0pt] at (13, 0, 0) {\includegraphics[width=3cm]{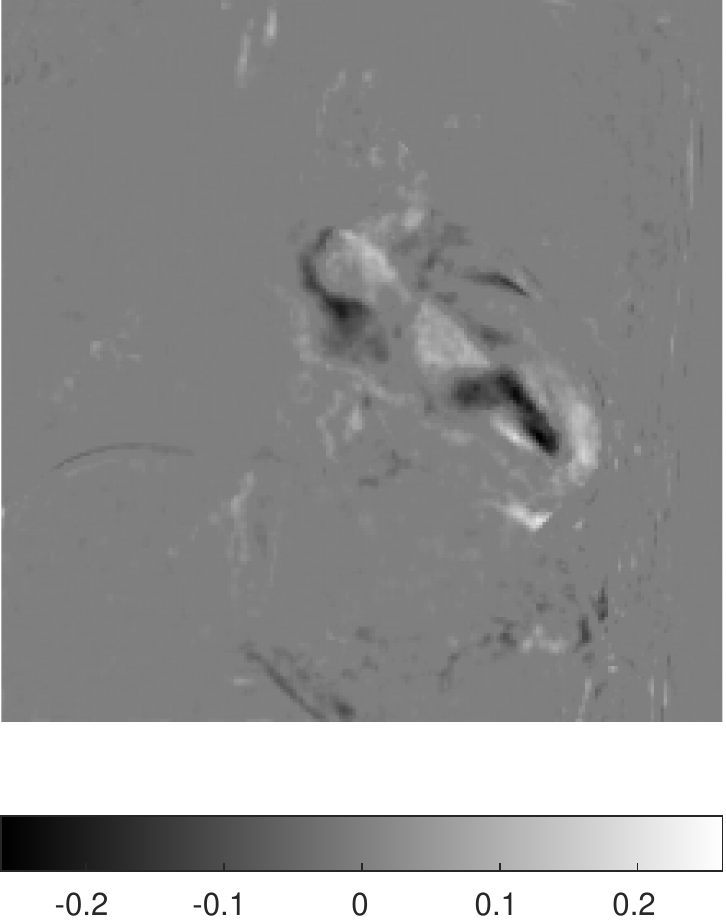}};
			\node[inner sep=0pt] at (16.25, 0) {\includegraphics[width=3cm]{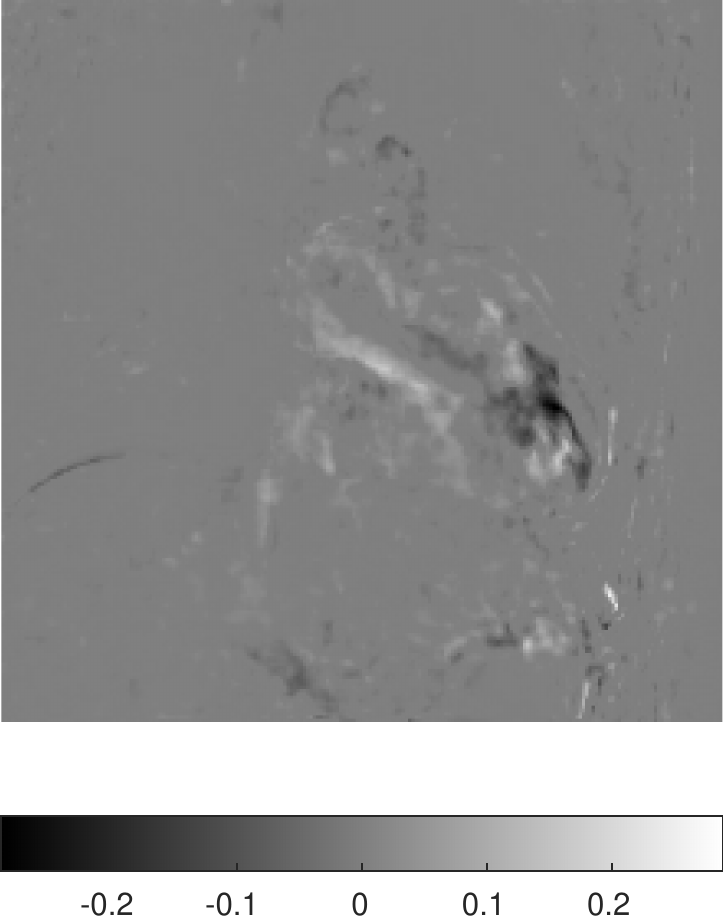}};
			
			\end{tikzpicture}
		}
		\caption{
			Sparse components $e_i = T_i(u^i) - l_i$ for all frames used in Fig.~\ref{fig:cardiac_results}.
			Nonzero entries are sparsely distributed across all images and primarily serve to correct the irregularities of the blood flow that were not captured by the low-rank components~$l_i$, i.e., that were not representable in a low-dimensional linear subspace.
			Thus, a meaningful low-rank approximation of the motion-corrected images is only enabled by allowing for sparsely distributed outliers
		}
		\label{fig:cardiac_sparse}
	\end{figure}

	Results of our approach are presented in \Cref{fig:cardiac_results,fig:cardiac_sv,fig:cardiac_sparse}.
	\Cref{fig:cardiac_results,fig:cardiac_sparse} show deformations, warped images and sparsity components for all example frames from Fig.~\ref{fig:cardiac_input}.
	In Fig.~\ref{fig:cardiac_sv}, singular values and singular vectors of $L - \bar{L}$ are analyzed in a presentation similar to Fig.~\ref{fig:ellipse_sv}.
	The quantitative evaluation in terms of landmark accuracy is visualized in Fig.~\ref{fig:cardiac_lm_accuracy}.
	
	Although results are generally more balanced than was the case for the synthetic data, our approach still outperforms the two competing methods on this real-world dataset in terms of landmark accuracy.
	We refer to Fig.~\ref{fig:cardiac_lm_accuracy}, where our method not only achieves the highest accuracy for the majority of the landmarks, but where it is the only out of the three methods that did not introduce additional motion to any landmarks in the registration process:
	both competing approaches generate several landmarks with worse accuracy than in the unregistered input sequence.
	
	In terms of the low-rank/sparse decomposition, we remark that our model was able to generate meaningful low-rank components alongside the actual motion-compensation:
	As seen from Fig.~\ref{fig:cardiac_sv}, the centered low-rank components matrix $L - \bar{L}$ is dominated by three singular value/vector pairs in which the singular vectors $s_1 - s_3$ exhibit clear visual congruences with the three considered phases of the heart cycle.
	Congruences thereby consist of highlighted anatomical structures and physiological features such as the mitral valve and the direction of blood flow.
	
	Furthermore, granting the method the ability to define sparse outlier components aided the generation of a meaningful low-rank approximation by filtering out highly irregular image feature such as the turbulences of blood flow (see Fig.~\ref{fig:cardiac_sparse}).
	
	\subsection{CDSOF}
	\label{subsec:res_cdsof}
	
	\begin{table}[b]
		\centering
		\renewcommand{\arraystretch}{1.25}
		\resizebox{.5\columnwidth}{!}{
			\begin{tabular}{| l | c c |}
				\hline
				& $D_{\dRPCA} + \TV$ & Nonlocal \cite{Sun2010}\\
				\hline
				``Blinking Arrow'' & 20m\,35s& 3m\,24s\\
				``Flying Snow'' & 20m\,36s & 3m\,46s\\
				``Shadow on Truck'' & 20m\,29s & 3m\,13s\\
				\hline
			\end{tabular}
		}
		\caption{Computation times of proposed approach and reference method for selected ``CDSOF''-sequences}
		\label{tab:times_cdsof}
	\end{table}
	
\begin{figure}[p]
		
		\centering
		\resizebox{.95\textwidth}{!}{
			
			\begin{tikzpicture}
			
			\node[above] at (0, 1.65) {\scriptsize ``Blinking Arrow'' (Frame \#10)};
			\node[above] at (5, 1.65) {\scriptsize ``Flying Snow'' (Frame \#1)};
			\node[above] at (10, 1.65) {\scriptsize ``Shadow on Truck'' (Frame \#9)};

			\node[inner sep=0pt] at (0, 0) {\includegraphics[width=4cm]{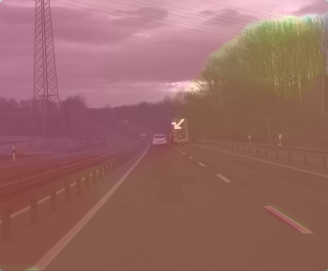}};
			\node[inner sep=0pt] at (0, -4) {\includegraphics[width=4cm]{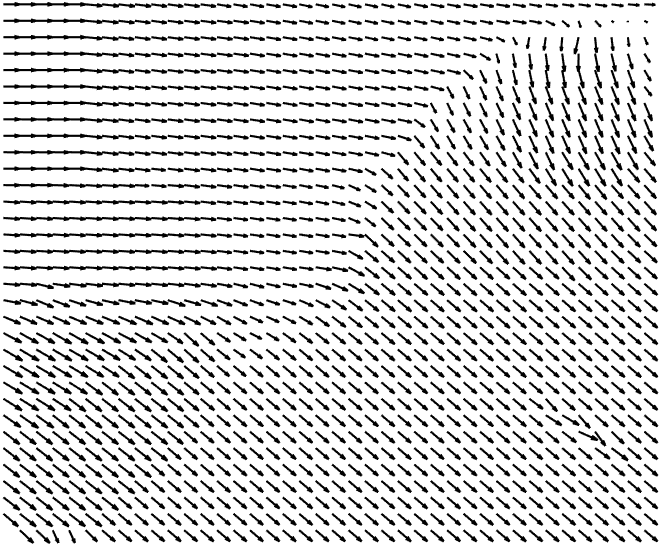}};
			
			\node[inner sep=0pt] at (5, 0) {\includegraphics[width=4cm]{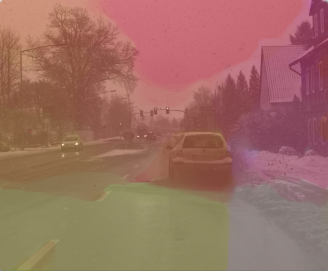}};
			\node[inner sep=0pt] at (5, -4) {\includegraphics[width=4cm]{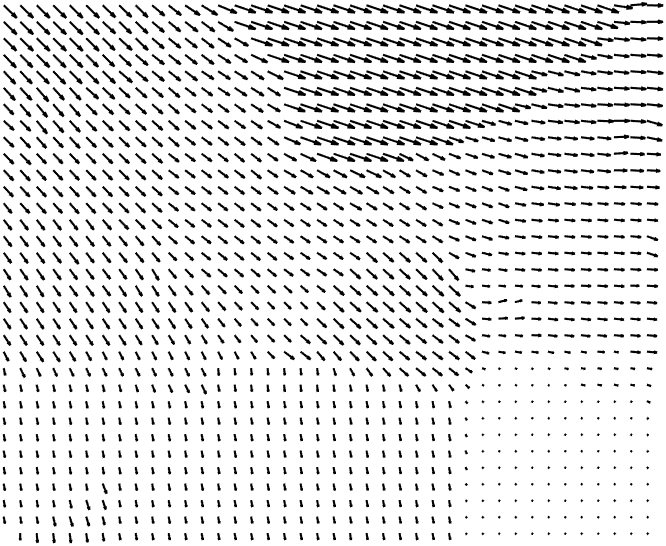}};
			
			\node[inner sep=0pt] at (10, 0) {\includegraphics[width=4cm]{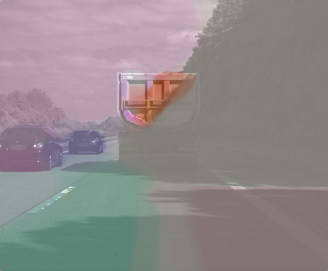}};
			\node[inner sep=0pt] at (10, -4) {\includegraphics[width=4cm]{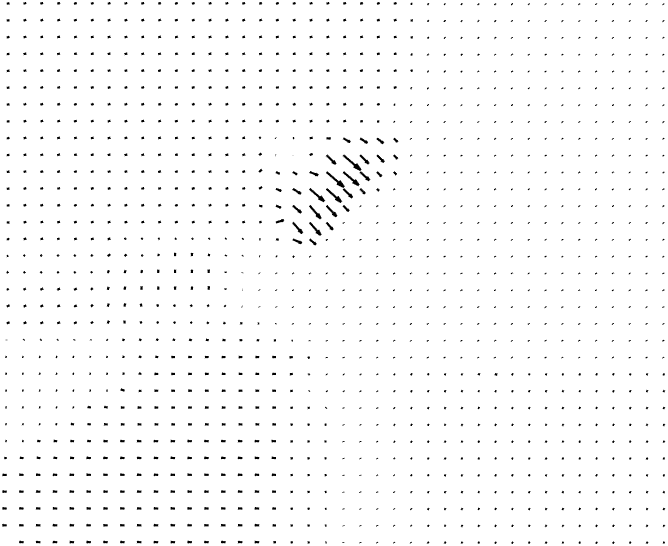}};
			
			\draw (-2.25, 1.75) -- (-2.5, 1.75) -- (-2.5, -5.75) node[midway, rotate=90, above] {$D_{\dRPCA} + \TV$ (proposed)} -- (-2.25, -5.75);
			\node[rotate = 90, above] at (-2, 0) {\scriptsize Color Coding};
			\node[rotate = 90, above] at (-2, -4) {\scriptsize Flow Field};

			\node[inner sep=0pt] at (0, -8) {\includegraphics[width=4cm]{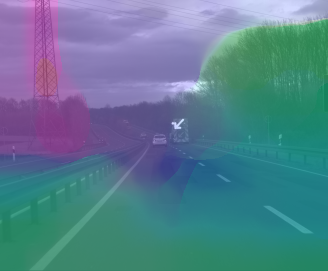}};
			\node[inner sep=0pt] at (0, -12) {\includegraphics[width=4cm]{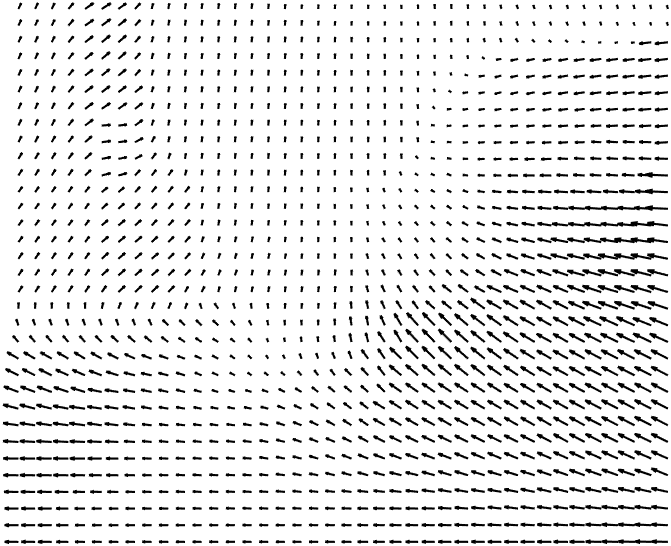}};
			
			\node[inner sep=0pt] at (5, -8) {\includegraphics[width=4cm]{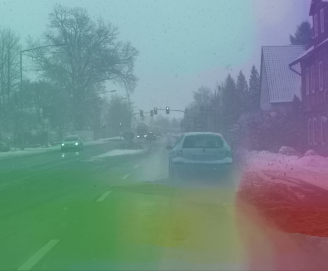}};
			\node[inner sep=0pt] at (5, -12) {\includegraphics[width=4cm]{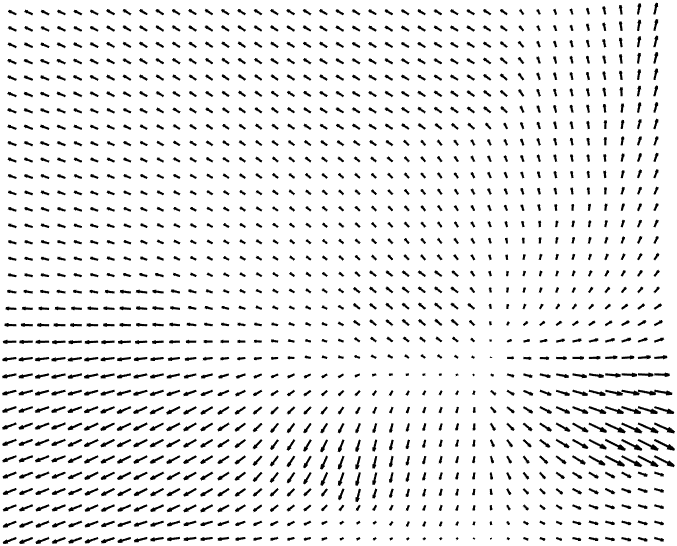}};
			
			\node[inner sep=0pt] at (10, -8) {\includegraphics[width=4cm]{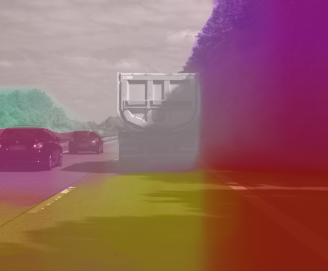}};
			\node[inner sep=0pt] at (10, -12) {\includegraphics[width=4cm]{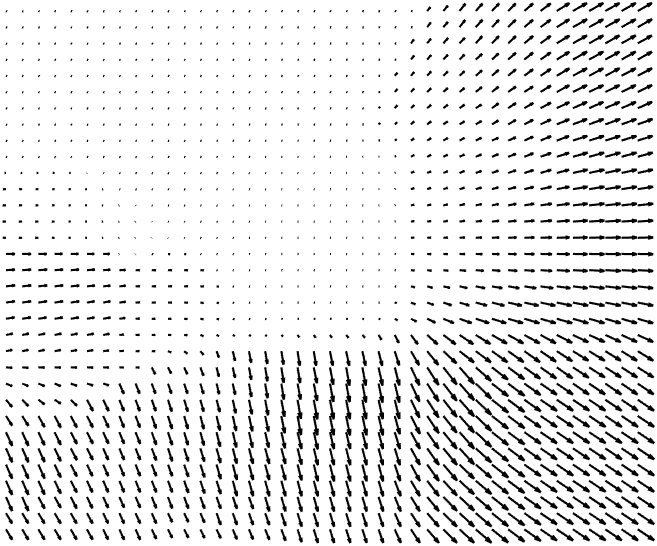}};
			
			\draw (-2.25, -6.25) -- (-2.5, -6.25) -- (-2.5, -13.75) node[midway, rotate=90, above] {Nonlocal \cite{Sun2010}} -- (-2.25, -13.75);
			\node[rotate = 90, above] at (-2, -8) {\scriptsize Color Coding};
			\node[rotate = 90, above] at (-2, -12) {\scriptsize Flow Field};
			
			\end{tikzpicture}
			
		}
		
		\caption{
			Comparison of optical flow estimation of the proposed model with the pairwise operating reference method from \cite{Sun2010}.
			For each sequence in Fig.~\ref{fig:cdsof_input}, one representative frame was selected, for which HSV color coded overlays and vector field presentations of the computed displacements are displayed.
			The ``Shadow on Truck''-sequence \textbf{(right column)} represents a clear failure case of our model as it was not able to distinguish the shadow casting from the actual physical motion (which the reference method succeeded in).
			For the other two sequences, our model was able to generate meaningful motion corrections.
			We again outline the difficulties involved in these sequences:
			The selected frame from the ``Blinking Arrow''-dataset \textbf{(left column)} features a lit traffic sign that is unlit in the reference frame (see Fig.~\ref{fig:cdsof_input}), while the ``Flying Snow''-dataset \textbf{(center column)} features heavy and irregular snowfall (compare the displayed frame to the reference in Fig.~\ref{fig:cdsof_input}).
			The apparent discrepancies between our approach and the method from \cite{Sun2010} are explained by the phenomenon discussed in subsection~\ref{subsec:res_cdsof}:
			Rather than aligning all deformable images with the fixed reference image (as the pairwise operating reference method does), our groupwise approach aligns these with another implicit reference.
			The explicit reference hence appears as an outlier in light of this implicit reference with misaligments being absorbed by the $\ell_1$-term of our distance metric~\eqref{eq:mod_measure}
		}
		\label{fig:cdsof_results}
		
	\end{figure}
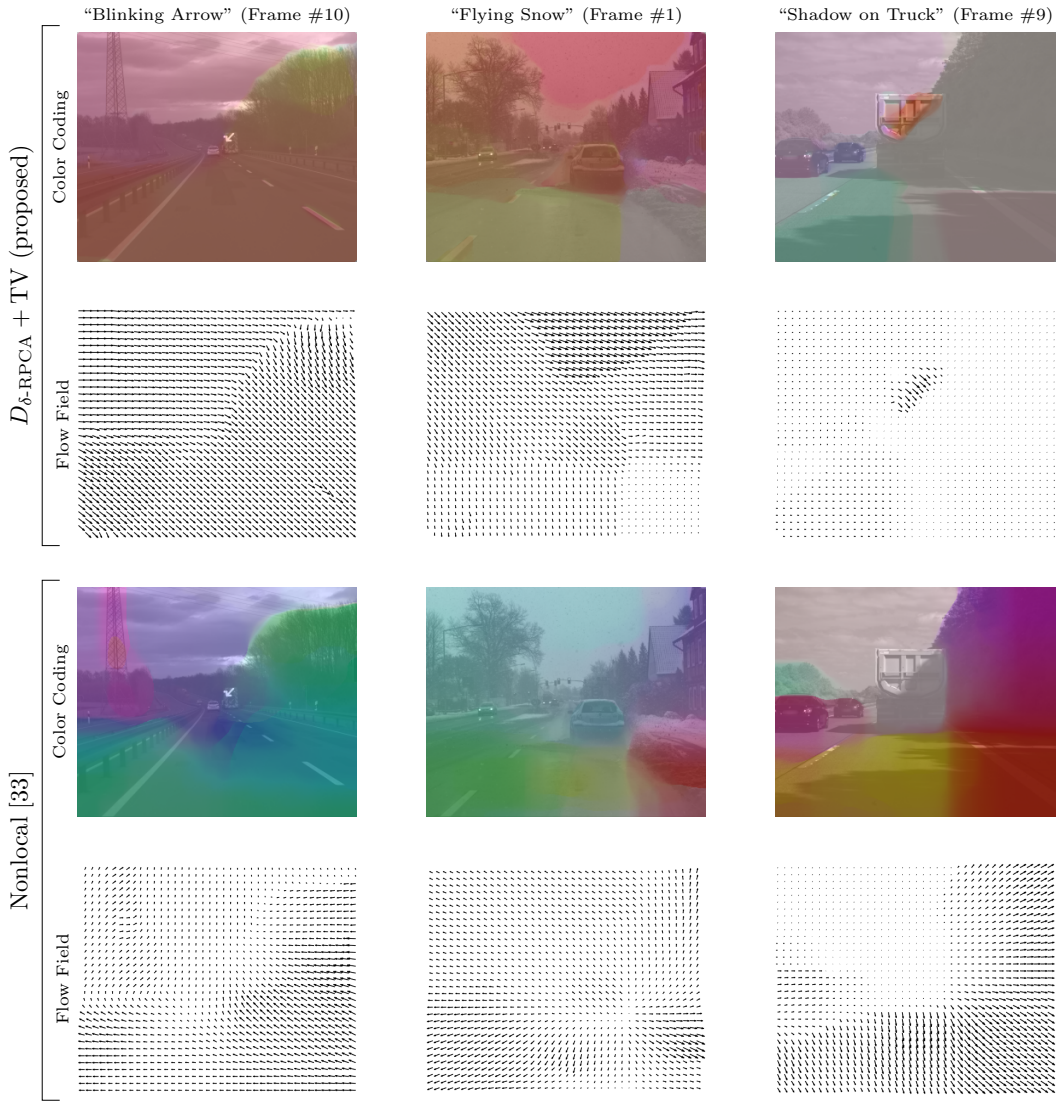

	A comparison of the motion estimation capabilities of our method with the pairwise operating reference \cite{Sun2010} for all sequences from Fig.~\ref{fig:cdsof_input} is given in Fig.~\ref{fig:cdsof_results}.
	
	In our model, we used the parameters $n_{lev} = 4$, $n_{iter}^1 = 16$, $ n_{iter}^j$ for $j \geq 2$, $\alpha = 0.91$ across all three examples as well as $\mu = 0.075$ for the ``Blinking Arrow'', $\mu = 0.045$ for the ``Flying Snow'' and $\mu = 0.125$ for the ``Shadow on Truck'' sequences.
	In the reference method implementation, we kept all parameters at standard values except for the regularization strengths, which were set to $\lambda = 10$ for the ``Blinking Arrow'' and to $\lambda = 20$ for both the ``Flying Snow'' and the ``Shadow on Truck'' sequence.
	
	Upon inspection of the results in Fig.~\ref{fig:cdsof_results} and Tab.~\ref{tab:times_cdsof}, it is clear that our approach was mostly outperformed by the reference method in terms of accuracy and computational efficiency.
	This is especially true for the sequence entitled ``Shadow on Truck'', where our method failed to produce a meaningful motion correction.
	For the other two sequences, our model was able to generate motion fields that successfully aligned all deformable (non-reference) images in the presence of disturbances such as snow flakes and blinking signs.
	
	We however point to the observation, that this alignment was not constructed with respect to the explicitly given reference but to another implicit reference generated by our algorithm.
	In light of this implicit reference, the explicit one hence appears as an outlier.
	This phenomenon also explains the large discrepancies observed in  Fig.~\ref{fig:cdsof_results} between the motion fields generated by our model and those generated by the reference method.
	
	We draw two conclusions from the experiments:
	\begin{enumerate}
		\item The proposed method exhibits a notable sensitivity towards the degree to which input data meets the model assumption of decomposability into structural low-rank components and sparse outlier components.
		If variations in object appearance are too irregular or if distortions are too large in scale (as in the ``Shadow on Truck''-sequence), the approach might fail to produce meaningful solutions.
		\item Imposing an explicit reference on a groupwise operating method cannot be expected to produce deformations that align all deformable images to that reference.
		On the contrary, deformable images might rather be aligned to a more suitable implicit reference.
		We primarily attribute this phenomenon to the small relative weight of one fixed reference when compared to the remaining group of $N - 1$ deformable images.
	\end{enumerate}
	
	We emphasize that the goal of this experiment is not to compete with a specialized method for a different domain (pairwise registration), but rather to give an indication of its usefulness on challenging non-medical real-world sequences.
	
	\section{Concluding Remarks}
	\label{sec:conclusion}
	In this work, we have investigated a novel dissimilarity metric for groupwise image registration tasks based on low-rank and sparse decompositions.
	The proposed metric corrects the major drawbacks that the established RPCA-image distance from \cite{Peng2010,Heber2014} exhibited in the experiments of Sec.~\ref{sec:rpca_distance}. 
	It is primarily suited for registering image data, that features objects with recurring changes in appearance and that can be represented in a low-dimensional linear subspace with potential sparse outliers.
	We especially emphasize the advantage in interpretability, when dealing with threshold constraints instead of weighted penalties.
	
	We further developed a first-order primal-dual optimization framework for solving non-pa\-ra\-met\-ric registration tasks using our metric in conjunction with TV regularization, which can easily be replaced by other regularization techniques suited for the individual application.
	
	Experimentally, we were able to show the superiority of our method when compared to two commonly used groupwise registration models.
	The experiments included both synthetic and real-world image data, that met the assumptions made by our model well.
	
	We further investigated the robustness of our model on a number of test sequences from the optical flow community, where we found that albeit it was outperformed by a highly optimized reference method, our model was able to correct motion in presence of distortions like snow flakes obstructing the view and illumination changes from blinking signs.
	As an interesting phenomenon, providing a reference to a groupwise method did not result in deformable images being aligned to the reference, but rather in the reference being treated as an outlier by our model.
	It remains to be investigated, to what extent this behavior is a trait of our particular model or of the general groupwise approach.
	
	Another avenue for future research is the application of our proposed decomposition model to tasks other than image registration.
	
	\begin{center}
		{\bf \small Acknowledgements}
	\end{center}
	\vspace{-.75em}
	{\small \hspace{\parindent} The authors thank Allen D. Elster (\texttt{MRIQuestions.com}) for kindly providing the cardiac cine study used in this article. The authors further acknowledge support through DFG grant LE 4064/1-1 ``Functional Lifting 2.0: Efficient Convexifications for Imaging and Vision'' and NVIDIA Corporation.}

	\bibliographystyle{abbrv}
	{\footnotesize \bibliography{DeformableGroupwiseImageRegistration_HaaseHeldmannLellmann}}
	
\end{document}